%% file: main.tex
\documentclass{ieeeaccess}
\usepackage{cite}
\usepackage{amsmath,amssymb,amsfonts}
\usepackage{algorithmic}
\usepackage{graphicx}
\usepackage{textcomp}
\def\BibTeX{{\rm B\kern-.05em{\sc i\kern-.025em b}\kern-.08em
    T\kern-.1667em\lower.7ex\hbox{E}\kern-.125emX}}
    
\input{text/custom_packages}
\input{text/custom_commands}

\begin{document}
\history{Date of publication xxxx 00, 0000, date of current version xxxx 00, 0000.}
\doi{10.1109/ACCESS.2022.DOI}

\title{Tensor-based Sequential Learning via Hankel Matrix Representation for Next Item Recommendations}
\author{\uppercase{Evgeny Frolov}\authorrefmark{1},
\uppercase{Ivan Oseledets\authorrefmark{1,2}}
}
\address[1]{Skolkovo Institute of Science and 
Technology, 121205, Moscow, Russia}
\address[2]{Artificial Intelligence Research Institute, 105064, Moscow, Russia}

\tfootnote{This work was supported by the Russian Science Foundation Grant No. 22-21-00911.}

\markboth
{E. Frolov, I. Oseledets: Tensor-based Sequential Learning via Hankel Matrix Representation for Next Item Recommendations}
{E. Frolov, I. Oseledets: Tensor-based Sequential Learning via Hankel Matrix Representation for Next Item Recommendations}

\corresp{Corresponding author: Evgeny Frolov (e-mail: evgeny.frolov@outlook.com).}

\begin{abstract}
Self-attentive transformer models have recently been shown to solve the next item recommendation task very efficiently. The learned attention weights capture sequential dynamics in user behavior and generalize well. Motivated by the special structure of learned parameter space, we question if it is possible to mimic it with an alternative and more lightweight approach. We develop a new tensor factorization-based model that ingrains the structural knowledge about sequential data within the learning process. We demonstrate how certain properties of a self-attention network can be reproduced with our approach based on special Hankel matrix representation. The resulting model has a shallow linear architecture and compares competitively to its neural counterpart.
\end{abstract}


\titlepgskip=-15pt

\maketitle

\input{text/intro}

\input{text/related}

\input{text/problem}

\input{text/attention}

\input{text/proposed}

\input{text/comparison}

\input{text/experiments}

\input{text/results}

\input{text/conclusion}

\bibliographystyle{unsrt}
\bibliography{main}

\begin{IEEEbiography}[{\includegraphics[width=1in,height=1.25in,trim={0.5cm 0 0.5cm 0},clip,keepaspectratio]{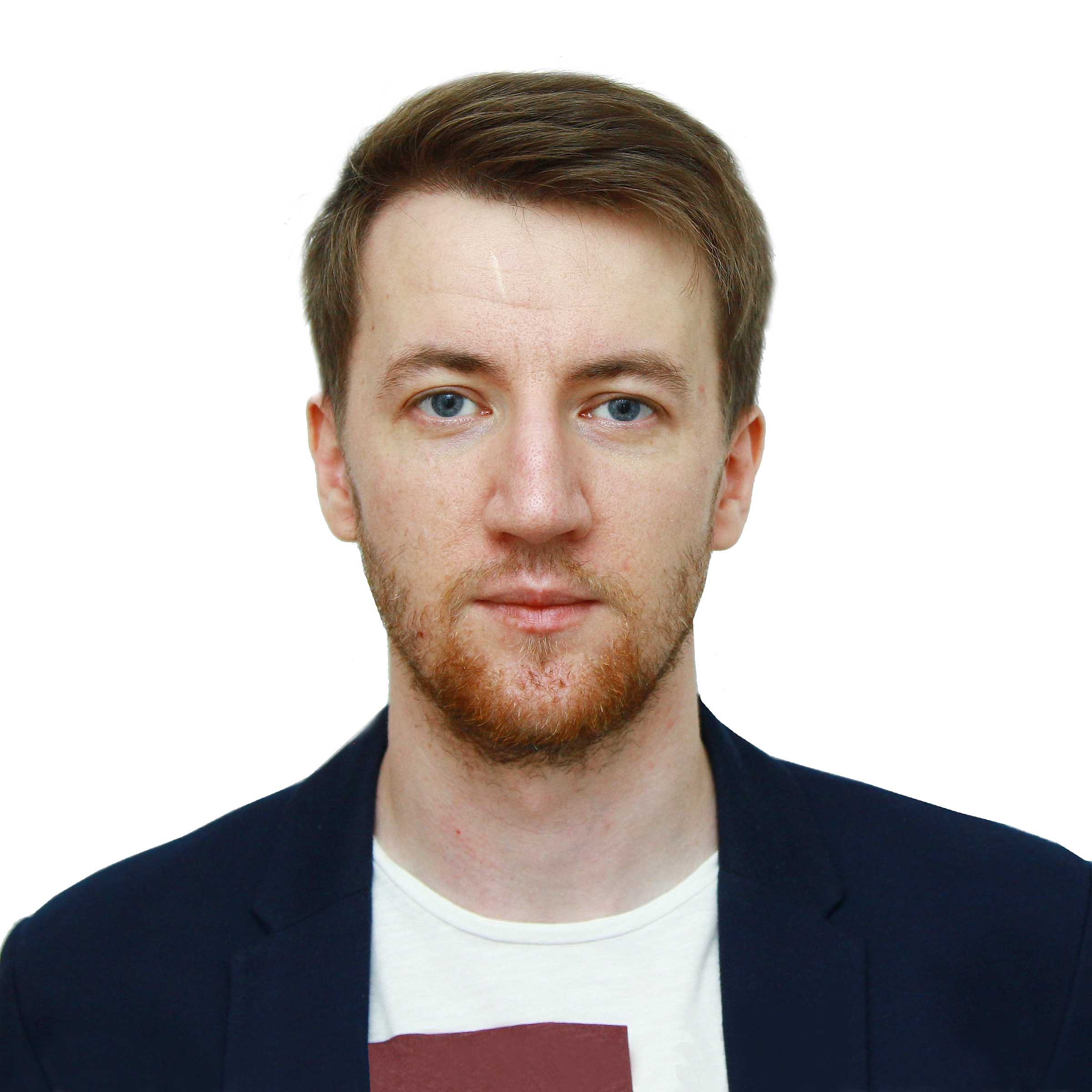}}]{Evgeny Frolov} graduated from Lomonosov Moscow State University in 2009. After graduation, he started developing career as an ICT industry professional, but returned to academia several years later. He receieved PhD degree in 2018 from Skolkovo Institute of Science and Technologies (Skoltech) under the supervision of Ivan Oseledets. Evgeny continues working at Skoltech as a research scientist, where he leads both academic and industrial research projects. His research is concentrated around building better bridges between practical challenges arising in the field of recommender systems and theoretical advances in relevant mathematical disciplines. He is especially attracted by algebraic and geometric methods, but also has broader interests. Some of Evgeny's work is published at the leading recommender systems conference ACM RecSys. He is also a co-author of a comprehensive survey on tensor methods in recommender systems.
\end{IEEEbiography}

\begin{IEEEbiography}[{\includegraphics[width=1in,height=1.25in,trim={21cm 0 3cm 2cm},clip,keepaspectratio]{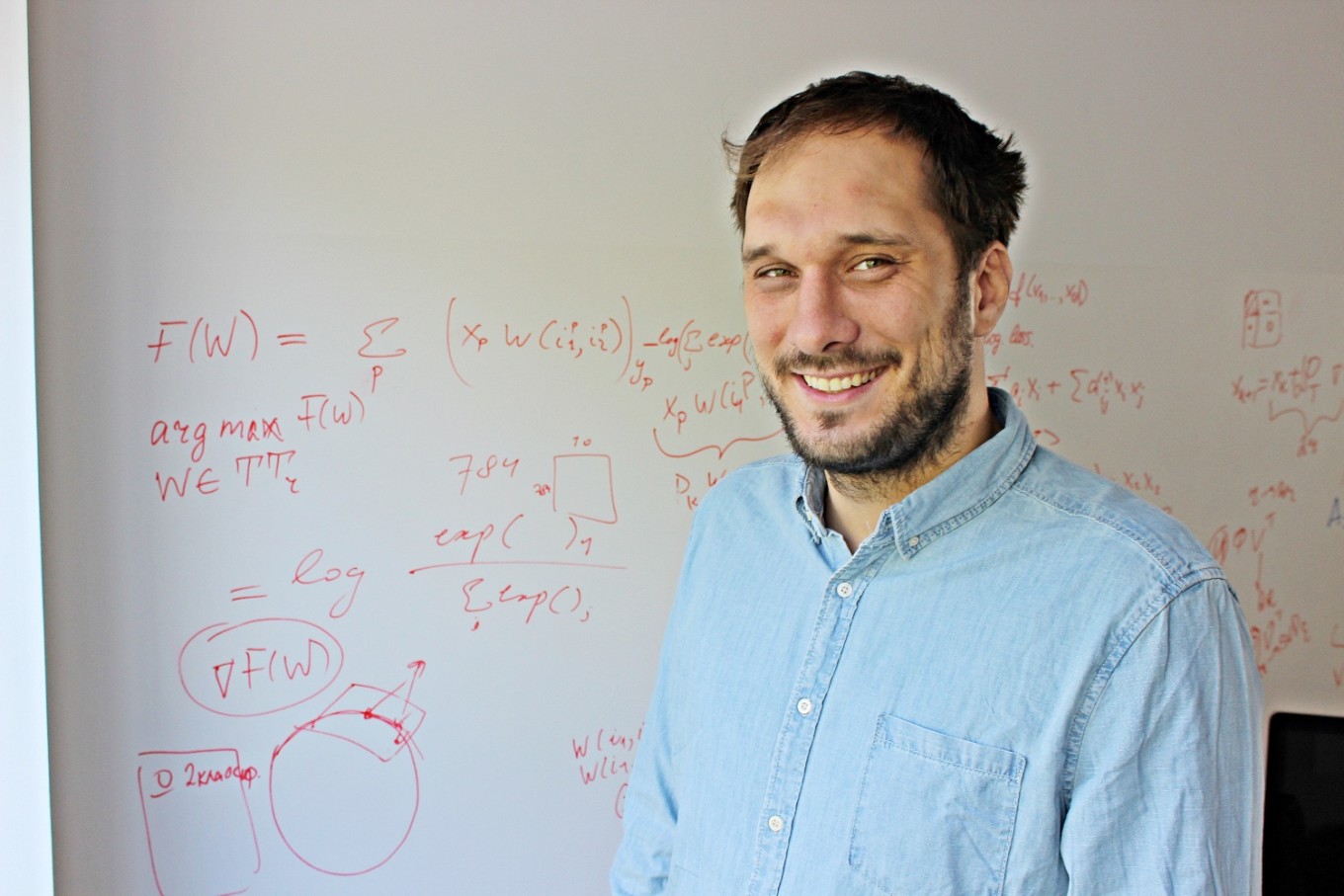}}]{Ivan Oseledets} graduated from Moscow Institute of Physics and Technology in 2006, got Candidate of Sciences degree in 2007, and Doctor of Sciences in 2012, both from Marchuk Institute of Numerical Mathematics of Russian Academy of Sciences. He joined Skoltech in 2013. 
Ivan's research covers a broad range of topics. He proposed a new decomposition of high-dimensional arrays (tensors) – tensor-train decomposition and developed many efficient algorithms for solving high-dimensional problems. His current research focuses on the development of new algorithms in machine learning and artificial intelligence, such as the construction of adversarial examples, theory of generative adversarial networks, and compression of neural networks. It resulted in publications in top computer science conferences such as ICML, NIPS, ICLR, CVPR, RecSys, ACL, and ICDM.
Professor Oseledets is an Associate Editor of SIAM Journal on Mathematics in Data Science, SIAM Journal on Scientific Computing, Advances in Computational Mathematics (Springer).
\end{IEEEbiography}

\EOD

\end{document}

%% file: text/custom_packages.tex
\usepackage{bm}
\usepackage[ruled]{algorithm2e}

%% file: text/custom_commands.tex
\newcommand{\tens}[1]{\mathbcal{#1}}
\newcommand{\matr}[1]{\mathbf{#1}}
\newcommand{\vect}[1]{\mathbf{#1}}
\newcommand{\inR}[1]{\in\mathbb{R}^{#1}}
\newcommand{\ind}[1]{{\left({#1}\right)}}

\newcommand{\argmax}{\mathop{\arg \max}\limits}
\newcommand{\frob}{\mathsf{F}}

\newcommand{\hankelop}[1]{\mathcal{H}\left[{#1}\right]}
\newcommand{\reserr}[2]{{#1}\pm{#2}}
\DeclareMathOperator{\ravel}{vec}
\DeclareMathOperator{\toprec}{toprec}
\DeclareMathAlphabet{\mathbcal}{OMS}{cmsy}{b}{n}

%% file: text/intro.tex
\section{Introduction}

 \PARstart{I}{n} recent years, the recommender systems (recsys) field witnessed a rapid development of new algorithms and their ubiquitous applications in real-world services. One of the pivotal moments was the announcement of the famous Netflix Prize competition that popularized the field. It intensified research in certain directions initially related to matrix and tensor factorization techniques, which were then overshadowed by the era of artificial neural networks (ANN). Along with greater formulation flexibility, the latter enjoyed massive development of advanced computational tools and convenient frameworks\footnote{e.g., PyTorch \mbox{https://pytorch.org} and Tensorflow \mbox{https://www.tensorflow.org}}, which increasing commoditization became vital for the practical success of ANNs.
 
 An additional boost was given by cross-disciplinary research, especially in the area of natural language processing (NLP) that historically served as the source of the practical and effective solutions, starting from the adaptation of the latent semantic indexing techniques for finding compact user and item representations to contemporary ANN-based models with sequential learning architectures.
Among the most recent developments in this direction, transformer architectures with self-attention blocks gained a lot of traction.
One of their remarkable features is the ability to efficiently and reliably extract patterns from sequential data.
Unsurprisingly, the sequential self-attention mechanisms found their use in the recsys tasks as well \cite{fang2020DeepSeq}.

One of the first successful adaptations was the Self-Attentive Sequential Recommendation model (SASRec) introduced in \cite{kang2018SARS}. It combined the idea of self-attention with an asymmetric weighting scheme respecting the causal order of items in a sequence. Unlike many previous neural network approaches that were shown to hardly compete with classical models in various scenarios \mbox{\cite{dacrema2021troubling,ludewig2021session,rendle2020mlpvsmf}}, SASRec was demonstrated to consistently outperform its non-neural competitors. It proved there was still a room for significant improvements in top-$n$ recommendations tasks despite the decades of competitive evolution of the field.

The latter fact led us to an intriguing question whether
\emph{it is still possible to get a comparable quality of recommendations without involving complex machinery of deep learning}. Is there still an undiscovered classical approach that, when properly tuned, would successfully compete with deep neural networks in the \emph{sequential learning} tasks the same way matrix factorization techniques do it in the standard collaborative filtering settings \cite{rendle2020mlpvsmf,dacrema2021troubling}? To answer this question, we aim to develop a non-neural approach that mimics the SASRec's self-attention component and embodies it into a shallow linear model. 
One of the best candidates for this endeavor and the closest predecessor to the neural networks era of recsys is tensor factorization, known to provide a great level of flexibility in various tasks \cite{frolov2017tensor,sidiropoulos2017tensor} and at the same time inducing only moderate overhead on the complexity of final solutions.
 
Hence, in this work, we propose a new sequence-aware tensor factorization approach to generate accurate next item recommendations. Our contributions can be summarized as follows:
\begin{itemize}
    \item We propose a special scheme for encoding sequential data in the tensor format and enrich it with a Hankel structure-based representation. The enriched representation exhibits a better capacity in mimicking attention mechanisms over the initial scheme. It enables a local short-range context for attention, which turns out to be an important factor for learning sequential patterns. We call it \emph{locally attentive} model in contrast to the initial \emph{globally attentive} one.
    \item We derive a new tensor factorization approach based on a generalized Tucker Decomposition. It mimics self-attention mechanism by imposing causal structure on an inner product space of the obtained sequential representation.
    \item We design an efficient ALS-based optimization technique that incorporates this attention mechanism into the learning process and accounts for a special sparse data format.
\end{itemize}

Our experiments demonstrate that the proposed non-neural tensor-based attention-mimicking approach successfully competes with SASRec in the next item recommendation task. To the best of our knowledge, \emph{this is the first attempt of finding a viable and more lightweight alternative to standard self-attention networks}. We believe that it can be further improved and extended in a way similar to how the SASRec model was improved upon since its first introduction in 2018 (see Section~\ref{sec:related-ann}). Moreover, the proposed sequential attention mechanism is not specific to tensor factorization and could be potentially adopted in neural networks as well. But it would require further exploration in this new direction, which is yet to be performed.

Hence, in this opening work, for the sake of more transparent and fair comparison, we match our approach only against the most straightforward implementation of sequential self-attentive learning, which the SASRec architecture conveniently provides. We leave further improvements and comparison with more recent and more sophisticated attention models for future work. We still hope, however, to bring into attention of the community a novel look on the problem of sequential learning, which may potentially lead to a new class of practical solutions.



%% file: text/related.tex
\section{Related work}
Two main areas of related research are considered in this section: 1)~sequential learning based on tensor data formats, and 2)~ANN-based sequential learning that utilizes self-attention mechanisms. Considering other forms of sequential learning, e.g., based on convolutional \cite{tang2018personalized} or recurrent \cite{hidasi2018GRUplus} neural networks, is out of scope of the current work. For the general overview of sequential learning in recommender systems, we refer the reader to \cite{wang2019sequential,fang2020DeepSeq}.
\subsection{Tensor-based Sequential Learning}
Many tensor-based approaches for capturing sequential patterns were proposed in the last decade. One of the first such approaches called Factorized Personalized Markov Chains (FPMC) \cite{rendle2010FPMC} was based on a (user, item, item) transition tensor obtained from statistics of which items are purchased after the current one within a single user session. FPMC used markovian principles to encode such transitions. The second and the third modes of the tensor encoded transition from and to an item correspondingly. The tensor was factorized with the help of the CP decomposition \cite{Kolda2009Tensors}.

Similar ideas, albeit with a bit more straightforward tensor-based approach, were explored in \cite{rettinger2012CARS}. The authors proposed to use a previously consumed item of a user directly as a contextual information for predicting the next one. In contrast to FPMC, no preliminary statistics calculation was necessary for constructing such tensor. The authors of \cite{hidasi2012FastCARS} also considered user's previously consumed items. However, instead of using previous items themselves they proposed to use item features as predictors to next user actions. They also generalized the approach to more than one previous item and considered two possibilities for encoding such information: either by assigning a separate new dimension to each of the previous item's features or by encoding all features into a single dimension. The former approach would probably capture more information from the interplay of different dimensions. However the resulting tensor would quickly become computationally intractable. Hence the authors opt for a simpler approach with a single dimension for all features. Worth noting that \cite{hidasi2012FastCARS} considered a general case of user sessions with repetitive actions. Hence, the recommendations were allowed to contain the items already consumed or known by a user. In this work, we exclude such repetitive consumption scenarios and \emph{require all recommended items to be new for a user}.

The ideas of encoding sequential information within some tensor representation were also explored in other domains, e.g., in NLP. The authors of \cite{cotterell2017TensorGram} provided a new interpretation of the word context in the standard tasks of semantic text analysis. They encoded positions of words that are within some distance from the current word in a sentence into a tensor format. The authors drew an analogy between such positional encoding and the sliding window of Skip-Gram models: both allow capturing semantic relations by looking at surrounding words. More recently, the authors of \cite{zhe2018EventTensor} proposed a sophisticated sequential learning model based on Hawkes process. All events preceding to the current moment were considered as potential sources that trigger the currently observed event. Previous events were assigned with different weights according to their recency via special triggering function. The approach was shown to be effective for events extraction and clustering.

The idea of using Hankel-structured matrices for the sequential data has a long history. One of the classic techniques that utilizes Hankel matrices representation within an algebraic framework is the singular spectrum analysis (SSA) \cite{Golyandina2013SSABook} sometimes also called the ``caterpillar approach''. It provides an effective solution for certain time-series completion and forecasting tasks \cite{agarwal2022mssa}. It even enjoys theoretical guaranties if the analyzed signal possesses certain harmonic properties. More recently, it was explored in application to image analysis \cite{yokota2020delayemb}. The authors build an image processing algorithm that expands image dimensions with the help of hankelization\footnote{here, we define hankelization as converting a vector form of sequential data into a corresponding Hankel matrix representation; this definition is different from the one given in the SSA-related works} (with additional padding and trimming of images). They show that the resulting methods are able to compete with a state-of-the-art deep neural network model. However, due to data characteristics, their algorithm operates on dense formats, which enables accelerated algebraic operations by e.g. Fast Fourier Transform (FFT). In our case, the characteristics of input data result in extremely sparse format, which makes direct application of standard fast computation techniques prohibitive. Moreover, no requirements on the order of pixels or their correlations is typically imposed and the learning algorithm is not designed for sequential data. We aim to address these challenges with our tensor-based formulation.


\subsection{Sequential Self-Attention in ANNs}\label{sec:related-ann}
There was a substantial progress in the development of ANN-based sequential learning models for recsys in recent years \cite{fang2020DeepSeq}. In the latest developments, various attention-based approaches prevailed other techniques, which can be explained by a great success of transformer models in the closely related NLP field. It has inspired many adaptations of successful transformer architectures to various recsys tasks. The already mentioned SASRec model \cite{kang2018SARS} is an excellent example of such a cross-disciplinary adaptation. Conveniently, it is also \emph{one of the most straightforward and concise implementations of sequential self-attention mechanism} for the next item prediction task. The model significantly outperforms non-neural sequential learning counterparts. Many of the most recent approaches \emph{use the same self-attention as a building block} within a more complex solution architecture.

For example, the authors of DUORec \cite{qiu2022duorec} propose to enhance self-attentive learning with additional contrastive regularization, which is aimed at resolving a representation degeneration problem. A popular BERT4Rec model \cite{sun2019bert4rec} adopts a BERT-like architecture from NLP with a special masking scheme of items in a sequence. It also utilizes a bi-directional attention approach in contrast to the uni-directional attention of SASRec. The S3Rec model \cite{zhou2020s3rec} employs mutual information maximization principle on top of the learned bi-directional attention to extract correlations within the observed sequences based on entities, their attributes, and even sequential segments.

There are also attempts to adopt attention networks as a general replacement for matrix factorization-based models. For example, the authors of the SeqFM model \cite{chen2020sequence} derive a special data representation for capturing sequential dynamics within the factorization machines framework \cite{rendle2010fm}. They also argue that modeling complex interactions between entities via simple aggregation of the corresponding dot-products in the latent space has a limited capacity. Hence, the authors propose to enrich an interaction modeling layer by replacing dot-products with self-attention blocks, which structure resembles a special form of the generalized matrix factorization \cite{he2017neural}.

All these examples present novel ideas that push state-of-the-art forward. However, as a consequence, the development of new sequential recommendation models becomes ``locked'' onto a specific paradigm of tackling the problem. The aforementioned ANN models build on top of the existing definition of the attention mechanism. With this work, we aim to look into an \emph{alternative paradigm with a different formulation of sequential attention}. We design a new sequential learning approach that combines the idea of Hankel matrix representation with a tensor-based encoding of item sequences. This formulation allows mimicking some properties of the regular sequential attention. Using economic sparse data formats, we build a new computational framework that implements this new type of sequential attention mechanism for the next item recommendation task.



%% file: text/problem.tex
\section{Sequence-Aware Tensor Factorization}\label{sec:attention_tensor}
We start by revisiting the problem of sequence-aware tensor factorization for the next item prediction. Our aim is to develop efficient computational scheme and use certain components of the SASRec architecture as an inspiration for our solution's design. We will gradually build our solution starting from the direct formulation of a third order tensor format for sequence-aware learning and then building up the final higher order solution with ``virtual'' dimensions expanded due to Hankel matrix representation.

\begin{figure}[b]
  \includegraphics[trim=0 85 0 85,clip,width=0.85\columnwidth]{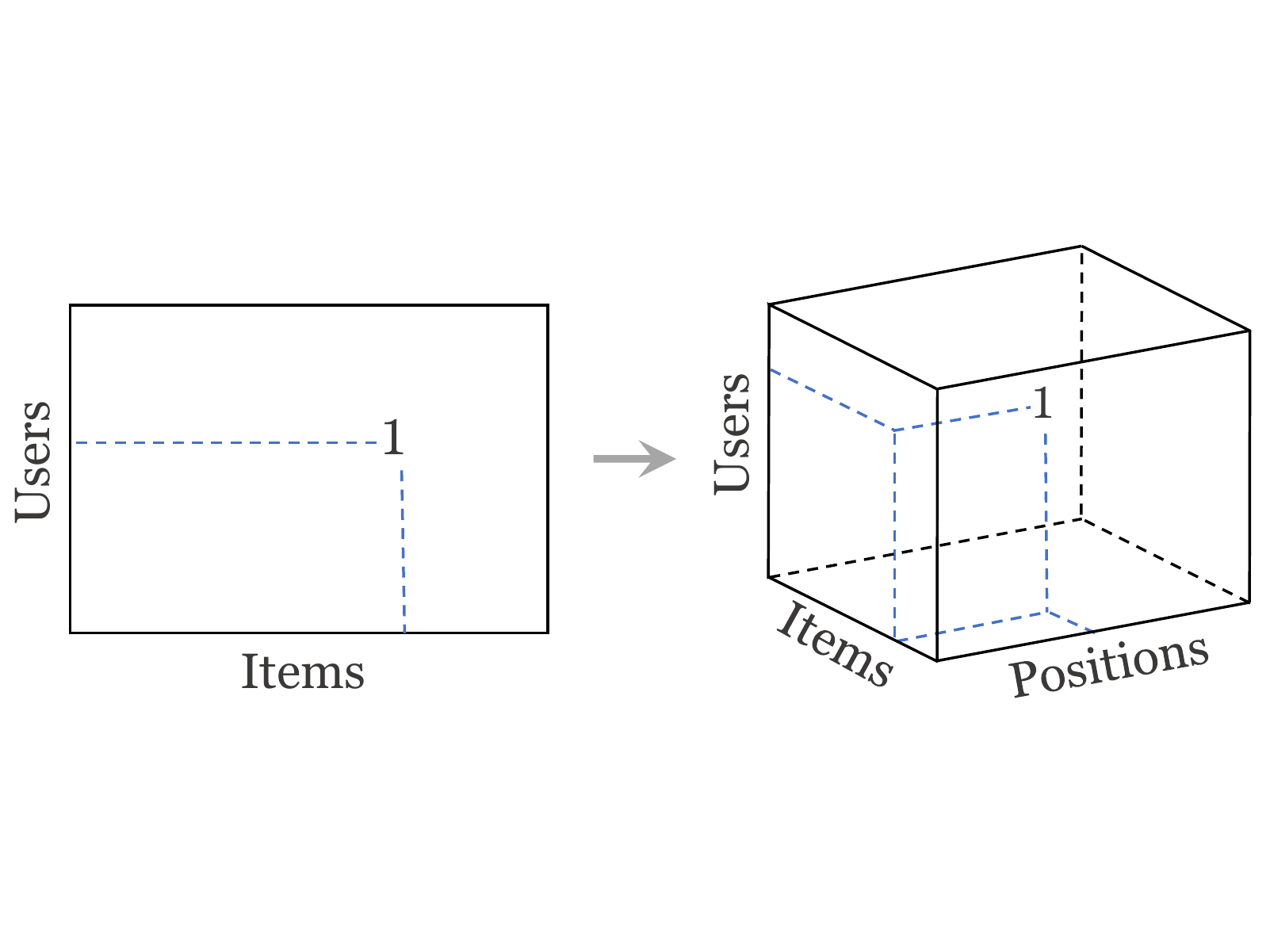}
  \caption{From simple interactions to positional information encoding in tensor format.}
  \label{fig:positional_tenor}
\end{figure}

\subsection{Problem Formulation}
Schematically, a third order tensor model can be formulated in terms of finding a scoring function $f_R$
\begin{equation}\label{eq:relevance}
    f_R: \text{User} \times \text{Item} \times \text{Position}\to\text{Relevance}
\end{equation}
that assigns some relevance score to each triplet of the observed interactions between users and items with respect to the positions of the latter in a user transactions history. The history length may vary as users exhibit different consumption behavior and may potentially become comparable to the size of entire item catalog. This would lead to a large dimension size for positional encoding and render certain computational challenges. However, it is natural to assume that only a relatively small number of the most recent items make a contribution into explaining current user decisions. Hence, one can truncate any user sequence to length $K$ with only the most recent items, where $K$ is much smaller than the catalog size $N$.

The described scheme expands a standard matrix of interactions \mbox{$\matr{X}=\left[x_{ij}\right]_{i,j=1}^{M,N}$} between $M$ users and $N$ items to a third order positional tensor $\tens{X}=\left[x_{ijk}\right]_{i,j,k=1}^{M,N,K}$. As users typically interact with only a small subset of all available items, the resulting tensor is extremely incomplete containing only a tiny fraction of known $x_{ijk}$ entries that correspond to the observed interactions (see illustration on Fig.~\ref{fig:positional_tenor}). There are various ways for handling such incomplete data, from simply ignoring all unknowns to assigning different weights depending on our confidence in the observations. In this work, we will use one of the most straightforward yet effective techniques proposed by the authors of PureSVD \cite{cremonesi2010PureSVD}, namely, zero-value imputation. Hence, the following binary tensor is formed:
\begin{equation}\label{eq:pos-tensor}
    \begin{cases}
        x_{ijk}=1 &\text{ if item }j\text{ is at position }p^i_k\text{ in user }i\text{ history,}\\
        x_{ijk}=0 &\text{ otherwise,}
    \end{cases}
\end{equation}
where $k=p^i_k-n_i+K$ and $n_i$ is the total number of items in the ordered history of user $i$. Any histories of length greater than $K$ are truncated so that $n_i\leq K,\;\forall i$. Note that by construction, the most recent item is always located at position $K$ independently of the length $n_i$ of the original sequence for any user $i$. If the number of items in a sequence is lower than $K$, padding with zeros is used to fill the remaining part up to the length $K$.

We note that padding with a special token other than zero is also possible. It will lead to a different semantics of items that were not yet interacted with versus a placeholder position in a sequence of incomplete length. It may potentially help capturing better sequential representations. For any observed user-item pair, it will make mode-3 tensor fibers\footnote{We use a common definition of tensor fibers, see \cite{Kolda2009Tensors} for details.} $x_{ij:}$ dense. Such padding would require special treatment for efficient computations. We opt for a simpler zero-padding representation and leave investigation of alternative padding schemes for future work.

Similarly to the matrix case, one can use tensor factorization techniques to compute embeddings for users, items, and positions and obtain a predictive next item recommendation model. Here, we focus specifically on the Tucker Decomposition (TD) format \cite{Kolda2009Tensors}, as it provides means for efficient generation of predictions by mere orthogonal projections through the learned latent space (explained in the next section). The learning objective is formulated similarly to the PureSVD case:
\begin{equation}\label{eq:objective}
    \|\tens{X}-\tens{R}\|^2_\frob\to\min,
\end{equation}
where $\|\cdot\|_\frob$ is a Frobenius norm. In the TD format, the objective can be optimized with the ALS-based higher order orthogonal iteration method (HOOI) \cite{deLathauwer2000tensorapprox}, which yields a low rank approximation $\tens{X}\approx\tens{R}$ of the following form:
\begin{equation}\label{eq:tucker}
    \tens{R} = \tens{G} \times_1\matr{U}\times_2\matr{V}\times_3\matr{W} \equiv [\![\tens{G};\;\matr{U}, \matr{V}, \matr{W}]\!].
\end{equation}
Here, $\times_n$ denotes an \emph{n-mode product} \cite{Kolda2009Tensors} between a tensor and a matrix; $\matr{U}\inR{M \times r_1}, \matr{V}\inR{N \times r_2}, \matr{W}\inR{K \times r_3}$ are \emph{columnwise orthonormal} matrices of embeddings corresponding to users, items, and positions. In most practical cases $r_1\ll\matr{M},r_2\ll\matr{N},r_3\ll\matr{K}$. Tensor $\tens{G} \inR{r_1 \times r_2 \times r_3}$ is called a core tensor of TD and a tuple of numbers ($r_1, r_2, r_3$) is called a multilinear rank.

\subsection{Next Item Prediction}\label{sec:tf-folding-in}
By relying on the orthogonality property of the factor matrices in TD, it is easy to derive a \emph{higher order analogy of the standard folding-in technique} \cite{furnas1988ir}. For our positional tensor it reads: 
\begin{equation}\label{eq:ho-folding-in}
    \matr{R}_\ind{i} = \matr{V}\matr{V}^\top\matr{P}_\ind{i}\,\matr{W}\matr{W}^\top,
\end{equation}
where $\matr{P}_\ind{i}$ is an $N \times K$ binary matrix with rows encoding items and columns encoding their positions in the user $i$'s history of actions. Correspondingly, the relevance matrix $\matr{R}_\ind{i}\inR{ N \times K}$ contains predicted scores for items with respect to their position in a sequence of length $K$.

Similarly to the matrix case \cite{cremonesi2010PureSVD}, we will use \eqref{eq:ho-folding-in} both for known users (i.e., when $\matr{P}_\ind{i}=\matr{X}_{i,:,:}$) and for warm-start users who were not present at the training phase as long as at least some of their historical preferences are known (so that matrix $\matr{P}_\ind{i}$ is not empty). Hence, we will omit the subscript $\ind{i}$ further in the text assuming that a matrix $\matr{P}$ provides sequential information on a subset of known items for some target user.

\begin{figure}[t]
  \centerline{\includegraphics[trim=0 70 0 60,clip,width=0.6\columnwidth]{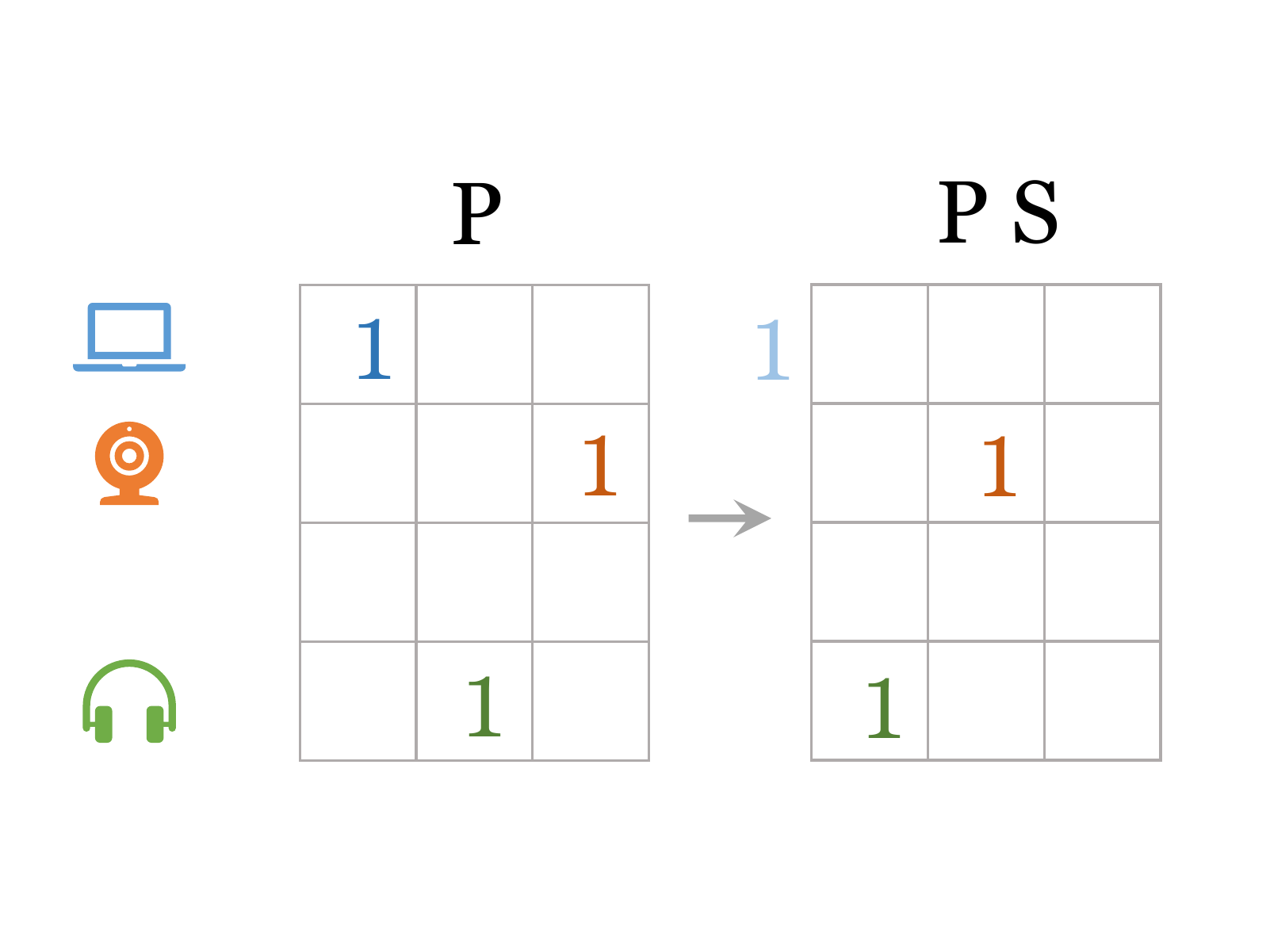}}
  \caption{Example of the shift operator acting on preferences matrix. All items positions are decreased by one, which leaves the last row empty.}
  \label{fig:shift}
\end{figure}

Conveniently, the structure of \eqref{eq:ho-folding-in} permits a straightforward rule of predicting the next item given a user's previous history. One just needs to shift all items in a user sequence one step left, which makes the last position in the sequence vacant. Then, applying \eqref{eq:ho-folding-in} to the shifted matrix $\matr{P}$ and taking the last column of the result (that corresponds to the last position in a sequence) will give us the relevance scores for the next item candidates.
Hence, given some user preferences matrix $\matr{P}$, the list of top-$n$ recommendations can be generated as:
\begin{equation}\label{eq:next-item}
    \toprec\left(\matr{P},n\right) = \argmax\limits^n \matr{V}\matr{V}^\top\matr{P}\matr{S}\,\matr{W}\vect{w}_K,
\end{equation}
where vector $\vect{w}_K$ is taken as the last row of $\matr{W}$, and \mbox{$\matr{S}=\left[\delta_{k,k'+1}\right]$} is a \mbox{$K\times K$} lower shift matrix that decreases positions of all observed items in $\matr{P}$ by one (see Fig.~\ref{fig:shift}). Note that if length of a user sequence is exactly $K$, the first item in the sequence gets discarded, which satisfies the \mbox{length-$K$} requirement for user histories in tensor construction.

Even though we added sequence-awareness into the tensor factorization model, it does not provide yet any attention mechanism. In the next section we describe how to bring attention on positions into the proposed tensor approach.

%% file: text/attention.tex
\section{Sequential Attention on Positions}
As demonstrated by the authors of SASRec, the weights of self-attention blocks exhibit on average a triangular structure with almost constant diagonals with respect to the position of items in a sequence (see \cite[Fig.~4]{kang2018SARS}). For the preceding positions, attention weights are either the same or lower than at the current position, but for any positions ahead the weights are all zero. This structure allows capturing causal patterns in the ordered sequences by assigning proper weights to sequence elements. The corresponding attention weights indicate that the model imposes correlations on items that are close to each other in a sequence in a strictly asymmetric order: no look-up into future items is allowed at any given moment of time. The next item may correlate with the preceding ones but not vice versa. Otherwise, it would provide an oracle hint during the training and impede the model's ability to accurately predict next items at the test phase.

\subsection{Mimicking Neural Self-Attention}
Unfortunately, the learning objective \eqref{eq:objective} with tensor defined as in \eqref{eq:pos-tensor} is not capable of capturing such sequential correlations.
Item positions in a sequence are encoded simply as a categorical variables and in this respect are no different from user and item encodings. \emph{There is no sense of sequential order or direction} in this representation. All positions will be treated by the model equally. Only at the prediction phase, we \emph{implicitly impose the order by focusing on the last position} for item scores prediction in \eqref{eq:next-item}.

However, with a slight modification to the tensor representation it is possible to add sequential correlations and mimic the positional attention mechanism of SASRec. Recall that self-attention in SASRec acts on a single user sequence. The latter corresponds to a matrix $\matr{P}$ of known user preferences. Positional correlations between items are then captured by the gram matrix $\matr{P}\matr{P}^\top$.

From here it immediately follows that \emph{any two items $j_1, j_2$ belonging to the same user are positionally uncorrelated}. The scalar product between their corresponding rows $\vect{p}_{j_1}$, $\vect{p}_{j_2}$ in matrix $\matr{P}$ will always be zero. For example, if matrix $\matr{P}$ belongs to a user $i$, then:
\begin{displaymath}
    \text{corr}_{\,\matr{P}}\!\left(j_1, j_2\right)\sim\!\left<{\vect{p}_{j_1}}, \vect{p}_{j_2}\right>\!=\!\sum_{k=1}^K x_{ij_1k}\,\cdot\,x_{ij_2k}\!=\!0\quad\forall i,\,j_1\neq j_2.
\end{displaymath}
Otherwise, it would mean that some items in a user history are located at the same position, which contradicts item enumeration in the sequence construction. We will call $\vect{p}_{j}$ \emph{positional vectors}.

\begin{figure}[t]
  \centerline{\includegraphics[trim=0 75 0 75,clip,width=.4\textwidth]{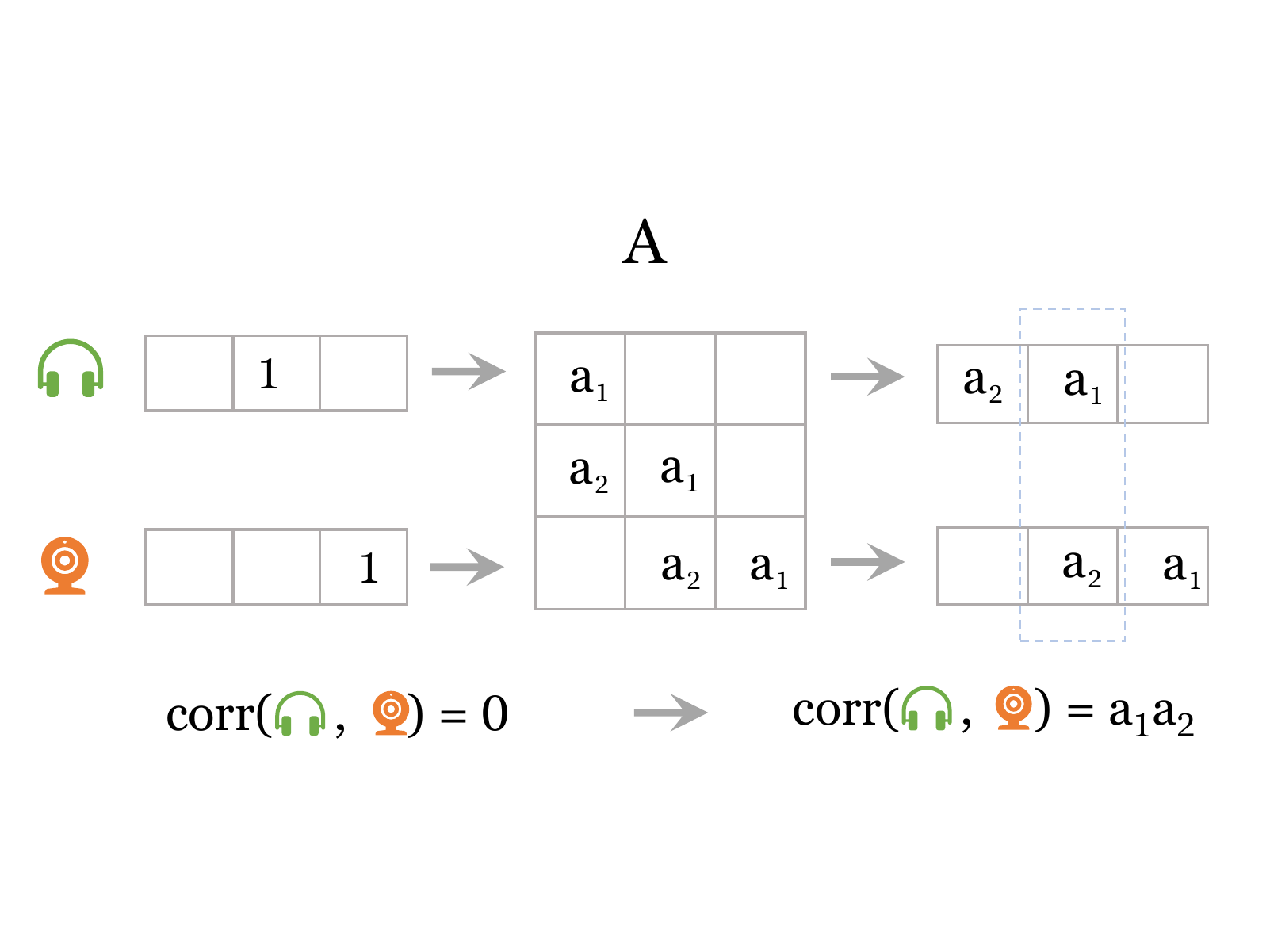}}
  \caption{Example of attention weights matrix acting on two positional vectors of length three. The scalar product between the vectors after applying $\matr{A}$ becomes non-zero.}
  \label{fig:attention}
\end{figure}

In order to add positional correlations and, therefore, mimic the self-attention mechanism, we replace the standard scalar product in the row-space of $\matr{P}$ with a bilinear form, which renders a new item correlation matrix:
\begin{equation}
    \matr{P}\matr{P}^\top \rightarrow \matr{P}\,\matr{C}\,\matr{P}^\top,
\end{equation}
where $\matr{C}$ is a $K\times K$ square symmetric matrix. By carefully crafting the structure of $\matr{C}$ one can impose additional relations within data. With the Cholesky Decompostion
\begin{equation}
    \matr{C} = \matr{A}\matr{A}^\top,
\end{equation}
where $\matr{A}$ is a \emph{lower triangular matrix}, it becomes easy to guess the appropriate for our task form. Inspired by the structure of self-attention weights generated by SASRec (i.e., \cite[Fig. 4]{kang2018SARS}), we require values along the main and lower-offset diagonals to be constant, which forms the following banded structure:
\begin{equation}
    \matr{A} = \begin{bmatrix}
    a_1 &  &  & \\
    a_2 & \ddots & \mbox{\Large 0} & \\
    \vdots & \ddots & \ddots & \\
    a_K & \dots & a_2 & a_1
    \end{bmatrix}.
\end{equation}
Each diagonal in $\matr{A}$ corresponds to an attention weight of a particular position in a sequence and the lower triangular structure imposes direction. An example of this matrix acting on the rows of $\matr{P}$ for the case of sequences of length $K=3$ is depicted on Fig.~\ref{fig:attention}.


Matrix $\matr{A}$ provides an asymmetric look-back mechanism by diffusing non-zero weights in a positional vector to preceding positions, which enables capturing directed positional correlations. The resulting effect of applying $\matr{A}$ in our approach is similar to what is achieved with triangular masking in the SASRec's self-attention. 
However, the weights in $\matr{A}$ are not learned by the model and we have to hand-craft their values. Probably, the simplest choice is \mbox{$a_k = k^{-f}$} for some factor \mbox{$f\ge0$}. When $f=0$, all preceding positions are equally important for an observation at the current position, and with $f>0$ items that are more distant in the sequence from the current one get lower attention. Other weighting schemes are also possible and can be designed based on domain knowledge or empirical assessment of the model. Learning the weights with the model instead of guessing them will require a different optimization scheme than the one proposed here (see the next section). We leave this question of end-to-end learning for future investigation.


\subsection{Tensor Factorization with Attention}\label{sec:gasatf}
If we were to solve just a 2D problem, we could simply use a generalized SVD formulation \cite{allen2014generalized,frolov2019hybridsvd} for handling bilinear forms instead of scalar products.
For the higher order case the scheme remains generally the same with a few modifications. The attention matrix must be applied across all users, and the $n$-mode product properties allow naturally achieving this simply by definition, which yields the following \emph{auxiliary tensor approximation task}:
\begin{equation}\label{eq:aux_approx}
    \left\|\tens{X}\times_3\matr{A}^\top-[\![\tens{G};\;\matr{U}, \matr{V}, \matr{W}]\!]\right\|^2_\frob\rightarrow\min_{\tens{G};\matr{U},\matr{V},\matr{W}}.
\end{equation}
It serves as a proxy for the task of approximating the original tensor $\tens{X}$. Conveniently, it can also be efficiently solved via the ALS-based HOOI algorithm.

Finding low-rank Tucker Decomposition of some $n$-dimensional tensor $\tens{Y}$ with the HOOI algorithm requires successive computation of truncated SVD of the compressed \mbox{tensor unfoldings}:
\begin{equation}\label{eq:hooi}
    \matr{Y}_k=\matr{Y}^\ind{k}\left(\matr{W}_n\otimes\dots\otimes\matr{W}_{k+1}\otimes\matr{W}_{k-1}\otimes\dots\otimes\matr{W}_1\right),
\end{equation}
where matrix $\matr{Y}^\ind{k}$ is the unfolding of $\tens{Y}$ along mode $k$ \cite{Kolda2009Tensors} and $\matr{W}_k$ are the sought factor matrices of the decomposition; $k=1,\dots,n$. In the case of positional tensor with attention, we have $\tens{Y}=\tens{X}\times_3\matr{A}^\top$, $n=3$, and $\matr{W}_k\in\{\matr{U},\matr{V},\matr{W}\}$.
The full learning procedure is listed in Algorithm~\ref{alg:hooi-3d}. 
For the reasons that will become clear in the next section, we call this model \emph{Globally Attentive Sequence-Aware Tensor Factorization} (GA-SATF).

\input{algos/hooi3d}

Note, we omit computations related to the core tensor $\tens{G}$. It remains unused in our implementation (as we emphasize in Section~\ref{sec:tf-folding-in}) and, unlike the classical implementation of the HOOI algorithm, is not used to track the learning progress and terminate ALS iterations. Thus, it can be safely ignored here\footnote{$\matr{U}$ factors are also unused in the model prediction, but are required during intermediate iteration steps, as outlined in Algorithm~\ref{alg:hooi-3d}.}. The stopping criterion in our case is defined by the growth of the target evaluation metric (see Section~\ref{sec:gridsearch} for more details).

Once the auxiliary tensor approximation task is solved, the original tensor is then approximated as \mbox{$\tens{X}\approx[\![\tens{G};\;\matr{U}, \matr{V}, \matr{\widehat W}]\!]$}, where the positional latent factors $\matr{\widehat W}$ of the original problem are obtained via:
\begin{equation}\label{eq:restore}
    \matr{\widehat W} = \matr{A}^{-\top}\matr{W}.
\end{equation}
There is no need to explicitly compute matrix inverse here. The task reduces to solving a triangular system of linear equations that can be efficiently performed due to the banded form of $\matr{A}$. Eq. \eqref{eq:restore} renders the following orthogonality property
\begin{equation}
    \matr{\widehat W}^\top\matr{C}\matr{\widehat W}=\matr{I},
\end{equation}
which indicates that the obtained latent space of $\matr{\widehat  W}$ is now enriched with positional attention correlations.

\subsection{Next Item Prediction with Attention}
Applying folding-in to the new model with attention yields a slightly different relevance prediction rule (c.f. \eqref{eq:ho-folding-in}):
\begin{equation}
    \matr{R} = \matr{V}\matr{V}^\top\matr{P}\matr{A}\matr{W}\left(\matr{A}^{-\top}\matr{W}\right)^\top,
\end{equation}
Based on \eqref{eq:restore}, we see that predictions now include positional latent spaces from both auxiliary and original problems. Correspondingly, shifting the preferences matrix and taking only the last position yields the following expression for the next item prediction task:
\begin{equation}\label{eq:next-item-new}
    \toprec_\text{GA-SATF}\left(\matr{P},n\right) = \argmax\limits^n \matr{V}\matr{V}^\top\matr{P}\matr{S}\,\matr{A}\matr{W}\vect{\hat w}_K,
\end{equation}
where $\vect{\hat w}_K$ corresponds to the $K$-th row of $\matr{\widehat W}$. By substituting \mbox{$\vect{p}=\matr{P}\,\matr{S}\,\matr{A}\,\matr{W}\vect{\hat w}_K$}, we arrive at the same form of top-$n$ recommendations as in PureSVD \cite{cremonesi2010PureSVD}, i.e.,  \mbox{$\toprec\left(\vect{p},n\right)=\argmax\limits^n\matr{V}\matr{V}^\top\vect{p}$}. Unlike the PureSVD case, $\vect{p}$ is not just an indicator of consumed items, but also carries directed sequential information on item correlations.





%% file: algos/hooi3d.tex
\begin{algorithm}[b]
    \SetKwInOut{Input}{Input}
    \SetKwInOut{Output}{Output}
    \Input{
        \quad Positional tensor $\tens{X}$ in sparse COO format.\\
        \quad Lower triangular attention matrix $\matr{A}\inR{K\times K}$.\\
        \quad Tensor decomposition ranks $r_1, r_2, r_3$.\\
    }
    \Output{\quad $\matr{U}, \matr{V}, \matr{W}$}
    Initialize $\matr{V}, \matr{W}$ as random matrices with orthonormal~cols.\\
    Compute $\matr{W}_A=\matr{A}\,\matr{W}$.\\
    \Repeat{stopping criteria met, see Section~\ref{sec:gridsearch}}{
        $\matr{U} \leftarrow$ $r_1$ leading left singular vectors of $\matr{X}^{(1)} \left(\matr{W}_A\otimes\matr{V}\right)$\\
        $\matr{V} \leftarrow$ $r_2$ leading left singular vectors of $\matr{X}^{(2)} \left(\matr{W}_A\otimes\matr{U}\right)$\\
        $\matr{W} \leftarrow$ $r_3$ leading left singular vectors of $\matr{A}^\top\matr{X}^{(3)}\left(\matr{V}\otimes\matr{U}\right)$\\
        $\matr{W}_A\leftarrow\matr{A}\,\matr{W}$\\
    }
    \caption{Globally Attentive Sequence-Aware TF}
    \label{alg:hooi-3d}
\end{algorithm}

%% file: text/proposed.tex
\section{Local Attention via Tensor Hankelization}
The base model introduced in Section~\ref{sec:attention_tensor} can be easily implemented. From preliminary experiments, we have identified that in some cases it provides a boost over non-sequential baselines. However, in other cases it failed to provide reasonable results and generally underperformed SASRec. We hypothesize that in an attempt to capture long-range patterns over the positional coordinate it looses a local context. Consider the following illustration.

Independently of the location in a sequence, at each moment of time (position in a sequence), a user's decision to consume the next item may be influenced by only a few preceding items. For example, purchasing a laptop may lead to the further purchase of a backpack. However, it does not make much sense to recommend a laptop to a user who just bought a backpack. Furthermore, going one step back, if prior to the laptop, the user also bought headphones, this purchase alone would not indicate that the user needed a backpack. We can say that seeing a laptop in the user's online order is a stronger predictor for the backpack purchase than seeing headphones there. Hence, to successfully predict the backpack purchase, a recommendation model must adequately discriminate between contributions of the headphones and the laptop by paying more attention to the latter.

The SASRec model naturally deals with this task due to its adaptive self-attention mechanism.
However, in the derived tensor representation with monotonically decaying attention weights $a_k$, the farther we look back into the sequence, the less distinctive become the corresponding positional attention weights. This, in turn, dissolves important sequential information. Closer to the beginning of a long sequence, the contribution of preceding items becomes almost equally important. Following the example above, the headphones and the laptop will get almost identical attention weights.

To overcome this problem and improve our models' capability of capturing localized in time effects, we design a sliding window representation that acts step by step on a sequence of items starting from the beginning. At each step, items within the window would represent a local decision context (most recent previous actions), while an entire user history would correspond to global user preferences. Consequently, instead of operating on the entire item sequence at once, the positional weighting is now applied within the sliding window to prevent attention dissolving and improve the discriminative power of the model.

\begin{figure}[b]
  \includegraphics[width=\columnwidth]{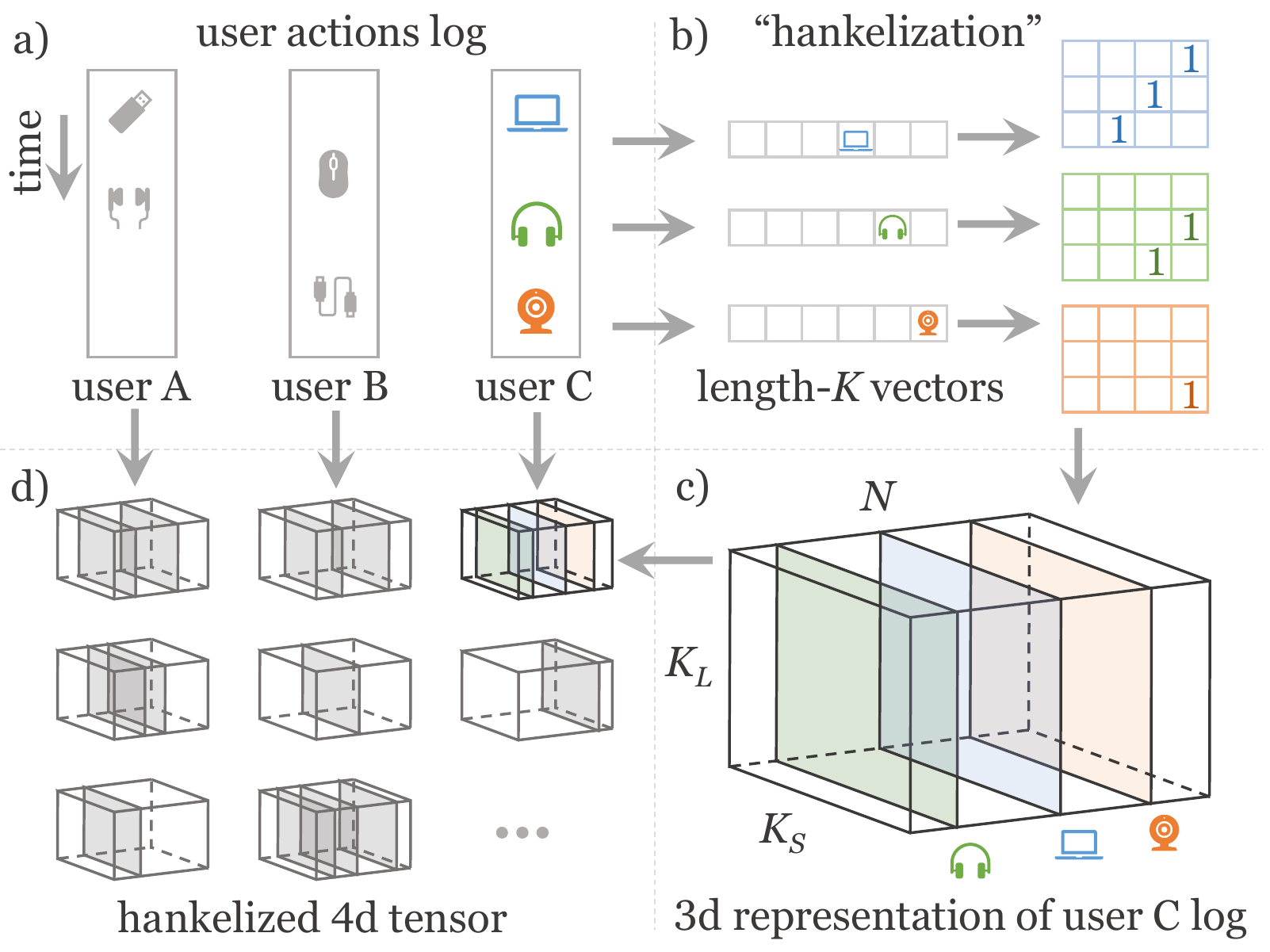}
  \caption{Constructing a 4d tensor that represents ordered transactions history: a) initial log data; b) convert positional vectors $\vect{p}_j$ of a user's $K$-most-recent items into their $K_L\times K_S$ Hankel matrix representation; c) the matrices (marked in color) become non-empty slices in a 3d tensor view of the user's history; d) all users are combined into a final tensor of size $M\times N\times K_L\times K_S$. Note that the new ``hankelized'' dimensions are never explicitly formed in actual computations and the original data is simply encoded in sparse COO format.
  }
  \label{fig:hankelization}
\end{figure}

Achieving this behavior \emph{requires changing sequential format}. Accordingly, we expand our third order tensor to four dimensions by \emph{transforming positional vectors encoded in the third mode into a Hankel matrix representation}: 
\begin{equation}\label{eq:hankelizator}
    \matr{X}_{(i,j)} = \hankelop{\vect{p}^\ind{i}_{j}},
\end{equation}
where $\vect{p}^\ind{i}_{j}$ is the positional vector of item $j$ in user $i$ history, $\hankelop{\,\cdot\,}$ is a linear operator that converts positional vectors of length $K$ into rectangular \mbox{$K_L\times K_S$} matrices with the Hankel structure, i.e., their skew diagonals are constant with values corresponding to the entries of an input vector; \mbox{$K_L+K_S-1=K$}. In signal processing tasks, $K_L \leq K/2$ is often considered a good choice. In that case it allows capturing harmonic signals. By varying values of $K_L$ one can recover harmonic signals of different periodicity.
In our case, we hope to extract useful sequential patterns and $K_L$ sets \emph{the local recency context size}. With $K_L = 1$ the model becomes equivalent to a standard third order tensor without attention.

By construction, any $\matr{X}_{(i,j)}$ always contains only one non-zero skew diagonal that corresponds to a position of item $j$ in a sequence of user $i$. Note that \emph{one can still operate on the same data without making any additional copies of it} by incorporating the described hankelization process into the calculations (see Section~\ref{sec:complexity}).
In that sense, the two new dimensions are completely ``virtual'' and are still encoded by the same positional vectors as before.
This new representation renders a fourth order tensor \mbox{$\tens{\widetilde X}\inR{M\times N\times K_L\times K_S}$} with ``virtual'' rear slices in the form of sparse Hankel-structured blocks: 
\begin{equation}
    \left(\tens{\widetilde X}\right)_{i,j,:,:}=\matr{X}_{(i,j)}.
\end{equation}
The tensor is depicted on Fig.~\ref{fig:hankelization}, where its non-empty slices $\matr{X}_{(i,j)}$ are marked with different colors.


\subsection{Locally Attentive Tensor Factorization}
As the positional vectors are now hankelized, we also have to redesign the attention mechanism. Recall, the goal of hankelization process is to capture local context within shorter parts of sequential data. Hence, the attention must be applied to the shortest dimension of the matrix $\matr{X}_{(i,j)}$, i.e., along the third mode of a fourth-order tensor $\tens{\widetilde X}$.
The optimization objective \eqref{eq:aux_approx} now transforms into:
\begin{equation}\label{eq:tucker4d}
    \left\|\tens{\widetilde X}\times_3\matr{A}_{L}^\top-[\![\tens{\widetilde G};\;\matr{U},\matr{V},\matr{W}_L,\matr{W}_S]\!]\right\|^2_\frob\rightarrow\min_{\tens{\widetilde G},\matr{U},\matr{V},\matr{W_L},\matr{W_S}}
\end{equation}
with \mbox{$\tens{\widetilde G} \inR{r_1 \times r_2 \times r_3\times r_4}$}, columnwise orthonormal factors $\matr{U}\inR{M \times r_1}$, \mbox{$\matr{V}\inR{N \times r_2}, \matr{W}_L\inR{K_L \times r_3}, \matr{W}_S\inR{K_S \times r_4}$}, and lower triangular attention matrix $\matr{A}_{L}$ of size $K_L\times K_L$. Instead of a single latent space $\matr{W}$ for positional embeddings we now have two subspaces. The original representation of the third mode (where the attention is applied) is recovered by $\matr{\widehat W}_L = \matr{A}_{L}^{-\top}\matr{W}_L$, which gives the approximation of $\tens{\widetilde X}$ as $\tens{\widetilde R}=[\![\tens{\widetilde G};\;\matr{U},\matr{V},\matr{\widehat W}_L,\matr{W}_S]\!]$.

The general TD-based learning scheme for the model remains similar to the one described in Section~\ref{sec:gasatf} with the corresponding substitutions \mbox{$\tens{Y}=\tens{\widetilde X}\times_3\matr{A}_{L}^\top$}, $n=4$, and \mbox{$\matr{W}_k\in\left\{\matr{U},\matr{V},\matr{W}_L,\matr{W}_S\right\}$}. The corresponding learning procedure is outlined in Algorithm~\ref{alg:hooi-4d}. As the attention weights are now applied within a local context window of size $K_L$, we call this model \emph{Locally Attentive Sequence-Aware Tensor Factorization} (LA-SATF).

\input{algos/hooi4d}

\subsection{Next Item Prediction with Local Attention}

Due to hankelization, the preference matrix $\matr{P}$ expands to the third order binary tensor $\tens{P}$ of size $N\times K_L\times K_S$. The corresponding higher order analogy of folding-in prescribes:
\begin{equation}
    \tens{R}_{L} = \tens{P}^{}_{L}\times_1\matr{V}\matr{V}^\top\times_2\matr{A}^{-\top}_{L}\matr{W}^{}_{L}\matr{W}_{L}^\top\matr{A}^{\top}_{L}\times_3\matr{W}^{}_{S}\matr{W}_{S}^\top.
\end{equation}
Subscript $L$ in $\tens{P}_L$ and $\tens{R}_L$ signifies that data was expanded due to the hankelization process with the local sequential context window of size $K_L$.
Similarly to \eqref{eq:next-item}, we are only interested in the item relevance scores predicted for the last position in a sequence, which correspond to the farthest vertical length-$N$ fiber of the \mbox{$N\times K_L\times K_S$} relevance prediction tensor $\tens{R}_{L}$.
Shifting user preferences to the left and using the Kronecker product properties yields:
\begin{equation}\label{eq:hankel-next-item}
    \toprec_\text{LA-SATF}(\matr{P}, n) = \argmax\limits^n \matr{V}\matr{V}^\top\vect{h}_{\matr{P}},
\end{equation}
where $\vect{h}_{\matr{P}}\inR{N}$ denotes hankelized sequential user preferences vector. Its elements are defined by
\begin{equation}
        \left(\vect{h}_{\matr{P}}\right)_j = \vect{\hat w}^\top_{K_L}\matr{W}_L^\top\matr{A}^\top_L\,\hankelop{\matr{S}^\top\vect{p}_{j}}\,\matr{W}_S\matr{w}_{K_S},
\end{equation}
with $\vect{\hat w}_{K_L}$ and $\vect{w}_{K_S}$ being the last rows of the matrices $\matr{\widehat W}_L$ and $\matr{W}_S$ respectively; $j=1,\dots,N$.




\subsection{Complexity analysis}\label{sec:complexity}
The main operation in the learning process of the model is computing truncated SVD of the compressed tensor unfoldings $\matr{Y}_k$ from \eqref{eq:hooi}.  
Hence, the complexity of the algorithm is defined by left and right matrix-vector products (matvecs) $\matr{Y}_k\vect{z}_k$ and $\matr{Y}^\top_k\vect{\bar z}_k$ with arbitrary dense vectors $\vect{z}_k$ and $\vect{\bar z}_k$ of conforming size. These matvecs are used for constructing Krylov subspace in the Lanczos procedure of truncated SVD. By using Kronecker product properties $\ravel{\left(\matr{AXB}\right)}=\left(\matr{B}^\top\otimes\matr{A}\right)\ravel(\matr{X})$, one can derive the corresponding matvec rules for the generalized HOOI. We split the computational complexity analysis into two parts: one corresponding to the coordinates of users and items, and another one describing the remaining ``virtual'' dimensions. For the sake of more transparent analysis, we assume \mbox{$r_1=r_2=d$}, and \mbox{$r_3=r_4=r$}. We also note that $r < d$ in all our experiments. 

\subsubsection{Main coordinates}
For the first two coordinates, corresponding to users and items, it is more convenient to define the rules in the element-wise manner. After a few algebraic simplifications, the final form of left matvecs with $\matr{Y}_1=\matr{\widetilde X}^{(1)}\left(\matr{W}_S\otimes\matr{W}_A\otimes\matr{V}\right)$ and $\matr{Y}_2=\matr{\widetilde X}^{(2)}\left(\matr{W}_S\otimes\matr{W}_A\otimes\matr{U}\right)$ reads:
\begin{equation}\label{eq:matvec12}
    \begin{gathered}
    \left(\matr{Y}_1\vect{z}_1\right)_i=\sum_{j}\ravel\left(\matr{W}_A^\top\,\matr{X}_\ind{i,j}\matr{W}_S\right)^\top\matr{Z}_1\,\vect{v}_{j},\\
    \left(\matr{Y}_2\vect{z}_2\right)_j=\sum_{i}\ravel\left(\matr{W}_A^\top\,\matr{X}_\ind{i,j}\matr{W}_S\right)^\top\matr{Z}_2\,\vect{u}_{i},
    \end{gathered}
\end{equation}
where $\vect{u}_i$ and $\vect{v}_j$ are the corresponding $i$-th and $j$-th rows of matrices $\matr{U}$ and $\matr{V}$, $\ravel\left(\matr{Z}_k\right)=\vect{z}_k$, and $\matr{W}_A = \matr{A}_L\matr{W}_L$.
The corresponding complexity of the terms under summation in \eqref{eq:matvec12} can be estimated as $O\left(r^2K\log K+r^2d\right)$.
The first term reflects computing $r^2$ entries of \mbox{$\matr{W}_\ind{i,j}=\matr{W}_A^\top\,\matr{X}_\ind{i,j}\matr{W}_S$}. The logarithm comes from the fact that, due to Hankel structure of $\matr{X}_\ind{i,j}$, entries of $\matr{W}_\ind{i,j}$ can be quickly computed via FFT between columns of $\matr{W}_A$ and $\matr{W}_S$.
The $r^2d$ term comes from multiplications from right to left.

The right matvec rules can be defined in a similar fashion via:
\begin{equation}\label{eq:rmatvec12}
    \begin{gathered}
    \matr{Y}^\top_1\vect{\bar z}_1=\sum_{i,j}\left(\left(\vect{\bar z}_1\right)_i\vect{v}_j\right)\otimes\ravel\left(\matr{W}_A^\top\,\matr{X}_\ind{i,j}\matr{W}_S\right),\\
    \matr{Y}^\top_2\vect{\bar z}_2=\sum_{i,j}\left(\left(\vect{\bar z}_2\right)_j\vect{u}_i\right)\otimes\ravel\left(\matr{W}_A^\top\,\matr{X}_\ind{i,j}\matr{W}_S\right),
    \end{gathered}
\end{equation}
where $\left(\vect{\bar z}_1\right)_i$ and $\left(\vect{\bar z}_2\right)_j$ are the $i$-th and $j$-th elements of $\vect{\bar z}_1$ and $\vect{\bar z}_2$ correspondingly. Calculating the terms under summations in \eqref{eq:rmatvec12} yields the same complexity $O\left(r^2K\log K+r^2d\right)$, where the $r^2d$ term now comes from the Kronecker product of two vectors.

Finally, for all matvecs in \eqref{eq:matvec12} and \eqref{eq:rmatvec12}, the result is non-zero only for the $(i,j)$ pairs corresponding to observed interactions. The total number of such interactions is bounded by $MK$, as we only encode sequences no longer than $K$ for each of $M$ users. The total complexity for this part becomes $O\left(MK(r^2K\log K+dr^2)\right)$. However, it does not account for the possibility of avoiding redundant calculations. Note, there are only $K$ distinct variants of $\matr{X}_\ind{i,j}$ independently of indices $\ind{i,j}$. Hence, the $K$ distinct variants of matrix $\matr{W}_\ind{i,j}$ can be pre-computed and cached before starting the main summation. The overall complexity is then reduced to $O\left((K^2r^2\log K+MKdr^2)\right)$ at the expense of storing $Kr^2$ additional elements in memory during calculations. With realistic values of $K$ and $r$, the memory overhead will be negligible comparing to the main storage of factor matrices. 


\subsubsection{Virtual dimensions}
For the remaining two ``virtual'' coordinates, matvec rules read:
\begin{equation}\label{eq:matvec34}
    \begin{gathered}
        \matr{Y}_3\vect{z}_3=\matr{A}_{L}^\top\sum_{i,j}\matr{X}_\ind{i,j}\,\matr{W}_S\,\matr{Z}_3\left(\vect{v}_j\otimes\vect{u}_i\right),\\
        \matr{Y}_4\vect{z}_4=\sum_{i,j}\matr{X}_\ind{i,j}^\top\matr{W}_A\,\matr{Z}_4\left(\vect{v}_j\otimes\vect{u}_i\right),
    \end{gathered}
\end{equation}
where matrices $\matr{Z}_k$ are such that $\ravel(\matr{Z}_k)=\vect{z}_k$.
Similarly, the complexity of the terms under summation in this case is estimated as $O\left(K_L+d^2r+Kr\right)$, where the $K_L$ term is attributed to the matvec between sparse Hankel matrix $\matr{X}_\ind{i,j}$ (or its transpose) containing at most $K_L$ non-zero elements, and a dense result of the right-to-left multiplications. We can ignore the $K_L$ contribution as it is subsumed by the $Kr$ term. Lastly, the corresponding right matvecs can be written as
\begin{equation}\label{eq:rmatvec34}
    \begin{gathered}
        \matr{Y}^\top_3\vect{\bar z}_3=\sum_{i,j}\left(\vect{v}_j\otimes\vect{u}_i\right)\otimes\left(\matr{W}_S^\top\matr{X}^\top_\ind{i,j}\matr{A}_L\vect{\bar z}_3\right),\\
        \matr{Y}^\top_4\vect{\bar z}_4=\sum_{i,j}\left(\vect{v}_j\otimes\vect{u}_i\right)\otimes\left(\matr{W}_A^\top\matr{X}_\ind{i,j}\vect{\bar z}_4\right).
    \end{gathered}
\end{equation}
These two matvecs add another $O\left(K_L^2+rK\log K+d^2r\right)$, where, as previously, the logarithm term comes from FFT-based calculations for constructing length-$r$ vectors of the form $\vect{w}_\ind{i,j}=\matr{W}^\top\matr{X}_\ind{i,j}\vect{z}$. The $K_L^2$ term appears due to $\matr{A}_L\vect{\bar z}_3$ product.

The overall complexity under summation in both \eqref{eq:matvec34} and \eqref{eq:rmatvec34} can be estimated as $O\left(K_L^2+rK\log K+d^2r\right)$, where we omit the $Kr$ term in favor of $rK\log K$.
The same pre-summation caching trick for $K$ distinct values of $\vect{w}_\ind{i,j}$ can be applied here, which results in the following estimate of the total complexity of this part: $O\left(rK^2\log K+MKd^2r\right)$. Note that the $K_L^2$ term is gone, as it is computed only once before caching, which makes its contribution negligible in comparison to the $rK^2\log K$ term.

\subsubsection{Comparison to SASRec}
As a final step, we need to combine contributions from all matvecs in \eqref{eq:matvec12}--\eqref{eq:rmatvec34}. Gathering all significant terms and omitting negligible ones gives the following estimate of the total complexity of a single iteration of HOOI: $O\left(K^2r^2\log K+MK\left(d^2r+dr^2\right)\right)$. An additional complexity is hidden in the Gram-Schmidt orthogonalization process that follows the Lanczos procedure in truncated SVD. However, its contribution, estimated as $O\left((M+N)d^2+Kr^2\right)$, can be disregarded after noticing that \mbox{$N\le MK$} and comparing it to the complexity of matvecs.

There is no direct way to perform a strict one-to-one comparison with the neural model as, for example, the embeddings size $d$ will have different optimal values for different classes of models. Moreover, the $r$ value is only present in the tensor-based model. Nevertheless, some rough estimates can be made. Assuming $r<d$, the LA-SATF's iteration complexity can be further simplified to $O\left(K^2r^2\log K+MKd^2r\right)$. Note that the major contribution comes from the second term as the number of users $M$ is significantly larger than any other factor in this estimation.  The complexity of each iteration of LA-SATF is thus \emph{either comparable to or slightly higher than the complexity of a single-epoch run of SASRec}, which is estimated as $O\left(MK(Kd+d^2)\right)$. Note, however, that ALS-based LA-SATF \emph{normally requires much fewer number of iterations to converge} in contrast to SGD-based SASRec. In all of our experiments, this number was around 4 or lower, while the optimal number of epochs in SASRec was at the order of 100.

The space complexity of our approach reduces to $O(Nd + Kr)$. We omit contribution of the tensor core $\tens{G}$ and the user embeddings $\matr{U}$, which would otherwise add $O(Md + d^2r^2)$. These factors are not required for generating recommendations and can be simply disregarded after the HOOI algorithm completes. Hence, in terms of space requirements, our model compares favorably to SASRec's $O(Nd+Kd)$.

%% file: algos/hooi4d.tex
\begin{algorithm}[t]
    \SetKwInOut{Input}{Input}
    \SetKwInOut{Output}{Output}
    \Input{
        \quad Positional tensor $\tens{X}$ in sparse COO format.\\
        \quad Local attention window size $K_L$.\\
        \quad Lower triangular local attention matrix $\matr{A}_L$.\\
        \quad Tensor decomposition ranks $r_1, r_2, r_3, r_4$.\\
    }
    \Output{\quad $\matr{U},\matr{V},\matr{W}_L,\matr{W}_S$}
    Initialize random $\matr{V},\matr{W}_L,\matr{W}_S$ matrices with orthonormal~cols.\\
    Compute $\matr{W}_A=\matr{A}_L\,\matr{W}_L$.\\
    Use hankelized tensor format $\tens{\widetilde X}$ (no data copying).\\
    \Repeat{stopping criteria met, see Section~\ref{sec:gridsearch}}{
        $\matr{U}\!\leftarrow$ $r_1$ leading left singular vectors of \hspace*{\fill}$\matr{\widetilde X}^{(1)}\left(\matr{W}_S\otimes\matr{W}_A\otimes\matr{V}\right)$\\
        $\matr{V}\!\leftarrow$ $r_2$ leading left singular vectors of \hspace*{\fill}$\matr{\widetilde X}^{(2)}\left(\matr{W}_S\otimes\matr{W}_A\otimes\matr{U}\right)$\\
        $\matr{W}_L\!\leftarrow$ $r_3$ leading left singular vectors of \hspace*{\fill}$\matr{A}_L^\top\matr{\widetilde X}^{(3)}\left(\matr{W}_S\otimes\matr{V}\otimes\matr{U}\right)$\\
        $\matr{W}_A\leftarrow\matr{A}_L\,\matr{W}_L$\\
        $\matr{W}_S\!\leftarrow$ $r_4$ leading left singular vectors of \hspace*{\fill}$\matr{\widetilde X}^{(4)}\left(\matr{W}_A\otimes\matr{V}\otimes\matr{U}\right)$\\
    }
    \caption{Locally Attentive Sequence-Aware TF}
    \label{alg:hooi-4d}
\end{algorithm}

%% file: text/comparison.tex
\section{Models comparison}
We compare a set of sequential and non-sequential models. Our main goal is to compare neural sequential self-attention model with our tensor-based sequential attention models. In addition to that, we provide a set of non-sequential baselines based on PureSVD \cite{cremonesi2010PureSVD} and its popularity-debiased variants known to perform better than the standard version. Note that we also apply the same debiasing trick to all tensor-based methods by default (more details on that are presented below). We additionally report scores for the most popular items recommendation model (MP).

\subsection{Neural Self-Attention on Sequences}
As we discussed in Section~\ref{sec:related-ann}, the main building block shared across many architectures of modern sequential learning models is the self-attention layer. All these models additionally implement various improvements on top of existing formulation of the self-attention mechanism. In contrast, our approach redefines the very mechanism of this attention. Hence, comparing our approach to the most recent developments in self-attentive sequential learning is unreasonable.

Even though it is potentially possible to incorporate the new type of attention into other models, it would make the feasibility analysis of our approach more difficult and convoluted. Hence, we seek for a more straightforward implementation of the self-attention to compare against. Apparently, the most reasonable candidate for such comparison is SASRec \cite{kang2018SARS}. It implements the essential parts of sequential self-attention almost in its purest form without additional tricks and extensions proposed in more recent models, which makes it a perfect target for comparison.

\subsection{SVD-based models}\label{sec:svd}
According to \cite{nikolakopoulos2019eigenrec}, the quality of SVD-based approaches can be significantly improved if an input data is properly normalized. We employ the technique proposed by the authors for both SVD-based and TD-based models. In the matrix case, given a binary matrix of observations $\matr{X}$, the corresponding normalized variant reads:
\begin{equation}\label{eq:norm_input}
    \matr{X}_\matr{D}=\matr{X}\,\matr{D},
\end{equation}
where $\matr{D}$ is an $N\times N$ diagonal matrix acting as a popularity debiasing factor. Its diagonal values are inversely proportional to the popularity of items and are calculated as $d_j=\left(\sum_i{x_{ij}}\right)^\frac{s-1}{2}$. The scaling factor $s$ serves as a hyper-parameter along with the rank of the decomposition. It allows adjusting the effect of popularity on the model learning. Higher values of $s$ put more emphasis on popular items, and they become more prevalent in recommendations. In contrast, lower values increase the overall sensitivity of the model to rare or niche items and help increasing recommendations diversity. Optimal values of this factor typically lie in the range $\left[-1, 1\right]$. After the model $\matr{X}_\matr{D} = \matr{U}\matr{\Sigma}\matr{V}^\top$ of rank $r$ is learned, the predictions are made via
\begin{equation}\label{eq:eigenrecproj}
    \toprec\left(\vect{p},n\right)=\argmax\limits^n\matr{V}\matr{V}^\top\vect{p},
\end{equation}
where \mbox{$\matr{V}\inR{N\times r}$}, and $\vect{p}$ is a length-$N$ binary vector of user preferences. 
We also note that the normalization scheme can be described in terms of the generalized SVD formulation that we used in the tensor models for incorporating attention. Hence, we can use the same original latent space restoration process as in \eqref{eq:restore} yielding a slightly different prediction rule:
\begin{equation}\label{eq:hybridproj}
    \toprec\left(\vect{p},n\right)=\argmax\limits^n\,\matr{D}^{-1}\matr{V}\matr{V}^\top\matr{D}\,\vect{p},
\end{equation}
In the experiments we treat the switch between regimes \eqref{eq:eigenrecproj} and \eqref{eq:hybridproj} as an additional hyper-parameter. Hence, we have only two implementations: standard PureSVD and PureSVD with normalized input \eqref{eq:norm_input}, which we call \emph{PureSVD-N}.

\subsection{Sequence-Aware Tensor Factorization with Attention}
We implement both models with the global \eqref{eq:next-item-new} and the local \eqref{eq:hankel-next-item} attention.
By default, we also use input data normalization as described above that acts on the frontal tensor slices. Hence, similarly to the matrix case, instead of the original tensors we approximate tensor $\tens{X}_\matr{D}=\tens{X}\times_2\matr{D}\times_3\matr{A}^\top$ in the GA-SATF case, and tensor $\tens{\widetilde X}_\matr{D}=\tens{\widetilde X}\times_2\matr{D}\times_3\matr{A}_{L}^\top$ in the LA-SATF case. Likewise, we can also apply two variants of the folding-in scheme depending on whether the original space is restored or not, i.e. the corresponding item space projector $\matr{VV}^\top$ in \eqref{eq:next-item-new} and \eqref{eq:hankel-next-item} is either replaced with $\matr{D}^{-1}\matr{V}\matr{V}^\top\matr{D}$ or used directly. We do not separately mark different models with additional labels like in the case of PureSVD and simply report the top-performing model assuming that the choice of item space projector is a model hyper-parameter.

%% file: text/experiments.tex
\section{Experiments}\label{sec:exper}

In this section we describe the general evaluation setup and preprocessing steps for performing experiments.  We take additional measures to comply with the fair comparison and rigorous evaluation requirements according to the best practices published in recent years. The source code to fully reproduce our work is openly published online\footnote{https://github.com/recspert/SATF.}.

\subsection{Evaluation methodology}
We generally follow the experimental setup described in the SASRec paper \cite{kang2018SARS}. However, following the recommendations and best practices from \cite{dacrema2021troubling}, we made two modifications related to how data is split into train and test parts, and how evaluation is performed.

Firstly, during the evaluation, we do not use item catalog subsampling to score against the true item hidden from the user history. This practice was shown to lead to inconsistent and unreliable results \cite{krichene2020sampled}. Hence, for each test user we predict scores on entire item catalog (excluding items previously seen by users) and then select top-$n$ items with the highest score to compare against the true item.

Secondly, we do not use simple leave-last-out procedure that hides items for evaluation based on just their position (i.e., the last item in a sequence). We use global timepoint-based splits instead. We define two time-intervals for validation and for final test, and split data accordingly from the end of a dataset. It helps addressing potential issues with oracle hints and ``recommendations from future'' \cite{meng2020datasplit,ji2020evaluation}, which may lead to unfair evaluation, especially in the case of sequential learning algorithms. The length of time intervals vary for different datasets due to different user activity. However, we ensure that each split contains approximately 5000 interactions. The resulting intervals are:  four months for each split in the ML-1M case, three weeks for each split in the AMZ-B case, and six weeks in the AMZ-G case. Finally, for Steam, we take two days for the test split and one day for validation. Hyper-parameter tuning is performed using the validation split. After an optimal configuration is found, we merge the validation part back into the training data, retrain the models with fixed hyper-parameters and perform final evaluation on the test split.

\subsection{Metrics}
In addition to HitRate (HR) and Normalized Discounted Cumulative Gain (NDCG) metrics reported in \cite{kang2018SARS}, we also report coverage (COV) measured as a fraction of the total number of unique items recommended by an algorithm to the total number of items in the training data. The latter metric serves as a proxy indicator of recommendations diversity. Lower values would indicate that an algorithm tends to focus more on some general patterns and does not offer high personalization. We do not group evaluation scores by users and perform calculations on a per-interaction basis. For example, the HR metric is calculated as the total number of hits (correct recommendations) divided by the total number of interactions in the test data. If a user appears several times in the test split, we combine the hidden items from the previous test interactions with the user's training history in order to predict the hidden item at the current step. The history is always time-sorted, which ensures the forward direction of these steps in time.

\subsection{Datasets}
We aim to repeat experiments conducted in the SASRec paper. Hence, we use the same four publicly available datasets that were analysed in the original paper: Movielens-1M (ML-1M), Amazon Beauty (AMZ-B), Amazon Toys and Games (AMZ-G), and Steam. However, as the data splitting procedure is different, we download datasets from their sources and perform preprocessing from scratch\footnote{Links to the datasets are included into the automated data processing pipeline in the data/prepare.py file in our repository}. We follow the same data preparation steps as in \cite{kang2018SARS}. We use 5-core filtering that leaves no less than five interactions per each user and each item. The explicit values of ratings are not used and are transformed into an implicit binary signal indicating the presence of a rating. Similarly to \cite{kang2018SARS}, the maximum allowed length of user sequences $K$ is set to $200$ for ML-1M and to $50$ on other datasets. We noticed that in the Steam dataset some users assigned more than one review to the same items.
We removed all such duplicate cases, which amounted to approximately 10\% reduction of the original dataset size.

The resulting statistics\footnote{We noted a discrepancy with the statistics provided in \cite{kang2018SARS} for both Amazon datasets. We were unable to identify the cause of it and provide both the entire code and links to datasets for fully reproducing our setup.} for the datasets are provided in Table~\ref{tab:datasets}. For each dataset, we also report average and median length of a user history (as a number of seen items), which serves as a hint for possible ranges of the local attention window sizes.

\input{tables/datasets}

\input{tables/main_results}

\subsection{Hyper-parameters grid search}\label{sec:gridsearch}
For the PureSVD-based models we tune the rank of SVD and the scaling factor $s$ described in Section~\ref{sec:svd}. As the SVD-based models are lightweight and quick to compute we fully explore a large grid of hyper-parameter values. For rank values $r$, their range is $\left(100,\ldots,3000\right)$  with step size gradually increasing from $100$ to $500$. For scaling $s$, the explored range is $\left(0.0,\ldots,1.0\right)$ with step size $0.2$.

For the tensor-based models, we explore values of $\left(r_1, r_2\right)$ of the multilinear rank in the range $\left(100,\ldots,1000\right)$ with step size $100$. In the case of the GA-SATF model, the positional mode rank $r_3$ takes values from $\left(5, 10, 15, 20\right)$. In the LA-SATF model, we have an additional hyper-parameter related to the attention window size $K_L$. The range of values for tuning $K_L$ depends on the dataset. It is estimated based on a median size of users' histories in a dataset (see Table~\ref{tab:datasets}). For example, in the ML-1M case, the set of allowed $K_L$ values is $\left(20, 40, 60, 80\right)$. For other datasets, $K_L$ takes values from $\left(1,2,5,10\right)$. Correspondingly, the positional ranks $\left(r_3, r_4\right)$ of the LA-SATF model take values from $\left(5, 10, 15, 20\right)$ in the ML-1M case, and from $\left(1, 2, 5, 10\right)$ for other datasets, excluding the values for which $r_3 \geq K_L$ or $r_4 \geq K_L$.
Finally, for all tensor-based models, the scaling factor $s$ takes values from $\left(0.0,0.2,0.4,0.6\right)$. We reduced this range after analyzing the performance of SVD-based models, where lower values of $s$ were consistently yielding better results.

In the case of SASRec model, we follow recommendations from the original paper but also vary values of the suggested hyper-parameters within reasonable ranges - batch size: $\left(64, 128, 256, 512\right)$, learning rate: $\left(0.00001, 0.0001, 0.001, 0.01\right)$, number of hidden units: $\left(64,128,256,512,728\right)$, number of attention blocks: $\left(1, 2, 3\right)$, and dropout rate: $\left(0.2, 0.4, 0.6\right)$. We use only one attention head like in the original paper, as adding more heads did not improve the results.

For all models, the target metric for optimal configuration selection is NDCG$@10$. Each model (except SVD-based) is allowed to explore $200$ grid points before termination. Optimal configurations found during the grid search are reported in our online repository. For all iterative algorithms (GA-SATF, LA-SATF, SASRec) we use an early stopping scheme based on the metric growth indicator. If the target metric ceases to improve within the last 3 evaluations, the iterations are stopped. Evaluation is performed after each iteration in the case of tensor-based model, and after every 20 epochs in the case of SASRec. The optimal number of iterations is stored and used for obtaining final test results reported in Section~{\ref{sec:results}}.


\subsection{Scalability comparison}\label{sec:scalability}
Performing a comprehensive and fair comparison of relative computational performance of different models is challenging in the absence of certain level of hardware and implementation compatibility. Current implementation of our tensor-based attention is CPU-based, whereas SASRec's implementation is based on PyTorch and hence is primarily optimized for running on GPU. Adapting our solution to GPU architectures is a viable next step, but it is out of scope of the current work due to a fair amount of technicalities to be addressed. Conversely, benchmarking only against the CPU-based runs of SASRec does not present a full picture. In order to make comparison as informative as possible in these challenging settings, we compare CPU runs of our tensor-based solution against both CPU- and GPU-based runs of SASRec.

There is also a need to address the problem of no exact match between the settings responsible for the number of learned parameters of the two models. Technically, one could introduce an ``effective'' dimension size, e.g., a fraction of the total number of learned parameters to the total number of items. However, such a measure will depend on a different set of hyper-parameters, which optimal values may significantly vary in different domains. For example, in the case of SASRec, it would be influenced by the number of attention blocks and the number of feedforward layers. On the other hand, in the LA-SATF model the effective dimension size would depend on the local attention window length and multilinear rank values. Attempting to take all these factors into account makes the scalability analysis cumbersome. With that in mind, \emph{we design two sets of experiments} that make this analysis more straightforward.


In the first set of experiments, we fix all the optimal hyper-parameters for both models except the \emph{items embedding size} $d$ for SASRec and \emph{the items latent space dimensionality $r_2$ of the multilinear rank} of LA-SATF. While these two hyper-parameters do not exactly correspond to each other, they are still both directly related to item representation, which is central to estimating the total number of learned parameters of a model and consequently its scalability. Hence, we gradually increase both $d$ and $r_2$ \emph{within the same range of values} and measure the time required to train each model. We repeat these measurements several times and report the averages. The results are presented in Fig.~\ref{fig:timings} and discussed in Section~\ref{sec:results}.

In the second set of experiments, we focus on measuring the general performance of all models with the best found configuration. Hence, we reuse all the optimal hyper-parameters as is and measure the total training time once again. Similarly, all measurements are performed several times and the average values are reported. These results can be found in Fig.~\ref{fig:optim_time} and are also discussed in Section~\ref{sec:results}.

\subsection{Hardware and implementation details}\label{sec:hardware}
The SASRec model was trained on a single NVidia Tesla V100 GPU using an open-source PyTorch implementation. The other models were trained on a CPU server with 64-core Intel(R) Xeon(R) CPU E5-2698 v3, 2.30GHz. LA-SATF and GA-SATF were implemented in Python and accelerated with Numpy and Numba. 






%% file: tables/datasets.tex
\begin{table}[t]
    \caption{Datasets statistics after pre-processing.}
    \label{tab:datasets}
    \begin{center}
    \begin{tabular}{llllll}
        \hline
        & & & \multicolumn{2}{c} {\#items per user:} &\\
        Dataset & \#users & \#items & average & median & density \\
        \hline
        Amazon Beauty & $22363$ & $12101$ & $8.9$ & $6$ & $0.07\%$ \\
        Amazon Games & $19412$ & $11924$ & $8.6$ & $6$ & $0.07\%$ \\
        Steam & $281205$ & $11961$ & $12.4$ & $8$ & $0.10\%$ \\
        MovieLens-1M & $6040$ & $3706$ & $165.6$ & $96$ & $4.47\%$ \\
        \hline
    \end{tabular}
    \end{center}
\vspace{-4mm}    
\end{table}

%% file: tables/main_results.tex
\begin{table*}[t]
\caption{Results of evaluation. All metrics are computed for top-$n$ recommendations with $n=10$.}
\label{tab:main_results}
\begin{center}
\begin{tabular}{llllllll}
\hline
 &       & \multicolumn{1}{c}{MP} & \multicolumn{1}{c}{PureSVD} & \multicolumn{1}{c}{PureSVD-N}    & \multicolumn{1}{c}{GA-SATF (ours)}   & \multicolumn{1}{c}{SASRec} & \multicolumn{1}{c}{LA-SATF (ours)} \\ \hline
NDCG & ML-1M & $\reserr{0.000}{0.000}$ & $\reserr{0.029}{0.002}$ & $\reserr{0.030}{0.002}$      & $\reserr{0.061}{0.002}$ & $\reserr{\underline{0.069}}{0.002}$           & $\reserr{\bm{0.072}}{0.003}$        \\
     & AMZ-B & $\reserr{0.002}{0.000}$ & $\reserr{0.046}{0.002}$ & $\reserr{0.047}{0.002}$      & $\reserr{0.043}{0.002}$ & $\reserr{\underline{0.055}}{0.003}$           & $\reserr{\bm{0.067}}{0.003}$        \\
     & AMZ-G & $\reserr{0.002}{0.000}$ & $\reserr{0.042}{0.002}$ & $\reserr{\bm{0.058}}{0.003}$ & $\reserr{0.046}{0.003}$ & $\reserr{\bm{0.055}}{0.003}$           & $\reserr{\underline{0.052}}{0.003}$             \\
     & Steam & $\reserr{0.000}{0.000}$ & $\reserr{0.020}{0.001}$ & $\reserr{0.043}{0.002}$      & $\reserr{0.007}{0.001}$ & $\reserr{\bm{0.060}}{0.002}$                  & $\reserr{\underline{0.047}}{0.002}$ \\ \cline{3-8}
HR   & ML-1M & $\reserr{0.000}{0.000}$ & $\reserr{0.060}{0.003}$ & $\reserr{0.061}{0.003}$      & $\reserr{0.112}{0.004}$ & $\reserr{\bm{0.134}}{0.004}$                  & $\reserr{\bm{0.132}}{0.004}$ \\
     & AMZ-B & $\reserr{0.004}{0.001}$ & $\reserr{0.082}{0.004}$ & $\reserr{0.087}{0.004}$      & $\reserr{0.079}{0.004}$ & $\reserr{\underline{0.100}}{0.004}$           & $\reserr{\bm{0.114}}{0.005}$        \\
     & AMZ-G & $\reserr{0.003}{0.001}$ & $\reserr{0.070}{0.004}$ & $\reserr{\bm{0.101}}{0.004}$ & $\reserr{0.074}{0.004}$ & $\reserr{\underline{0.094}}{0.004}$           & $\reserr{\underline{0.092}}{0.004}$             \\
     & Steam & $\reserr{0.000}{0.000}$ & $\reserr{0.039}{0.002}$ & $\reserr{0.084}{0.003}$      & $\reserr{0.013}{0.001}$ & $\reserr{\bm{0.115}}{0.004}$                  & $\reserr{\underline{0.091}}{0.003}$ \\ \cline{3-8}
COV  & ML-1M & $0.038$                 & $0.187$                 & $0.275$                      & $0.288$                 & $\underline{0.503}$ & $\bm{0.511}$                        \\
     & AMZ-B & $0.007$                 & $0.251$                 & $0.615$                      & $0.182$                 & $\bm{0.611}$ & $\underline{0.608}$                 \\
     & AMZ-G & $0.008$                 & $0.467$                 & $\underline{0.631}$          & $0.241$                 & $\bm{0.700}$ & $0.426$                             \\
     & Steam & $0.018$                 & $0.070$                 & $\bm{0.438}$                 & $0.047$                 & $0.080$                                       & $\underline{0.368}$                 \\ \hline
\end{tabular}
\end{center}
\vspace{-4mm}
\end{table*}

%% file: text/results.tex
\section{Results and Discussion}\label{sec:results}
The main results of the experiments are provided in the Table~\ref{tab:main_results}. The best results are in bold font, the second best are underlined. We report averaged scores and confidence intervals (except for the coverage as it is a single measurement and no error estimation is possible). If results of two models lay within their confidence intervals, we mark them the same way. For example, the difference in HR metric on the ML-1M dataset between SASRec and LA-SATF models is not significant, so we mark both as top scores.

\begin{figure*}[t]
  \includegraphics[width=\textwidth]{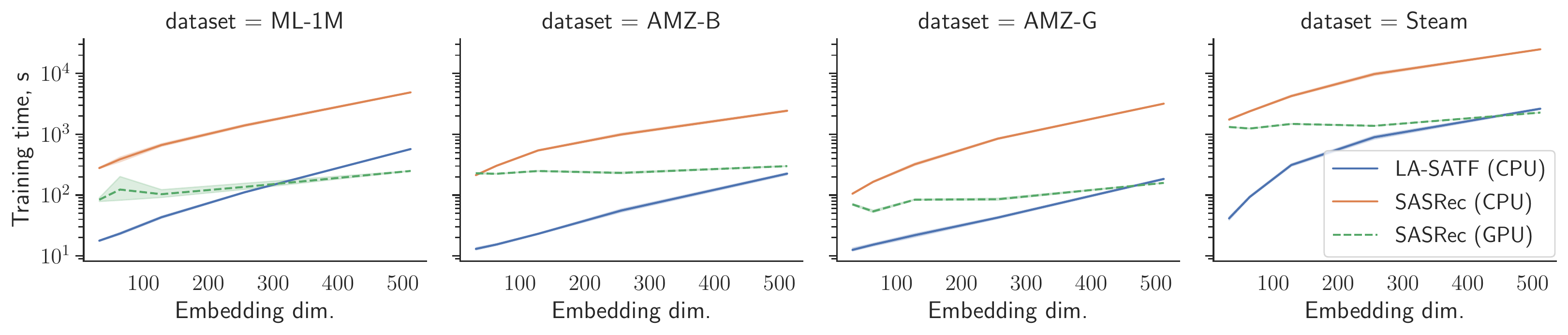}
  \caption{Model training time versus the number of learned model parameters. The latter is estimated via the item embedding dimension size ($x$-axis). The embedding dimension size corresponds to the value of $d$ in the case of SASRec and $r_2$ in the case of LA-SATF. Note, there are two types of measurements for the SASRec model: one CPU-based and another one GPU-based (marked with a dashed line).
  }
  \label{fig:timings}
\end{figure*}

\begin{figure}[b]
  \includegraphics[width=\columnwidth]{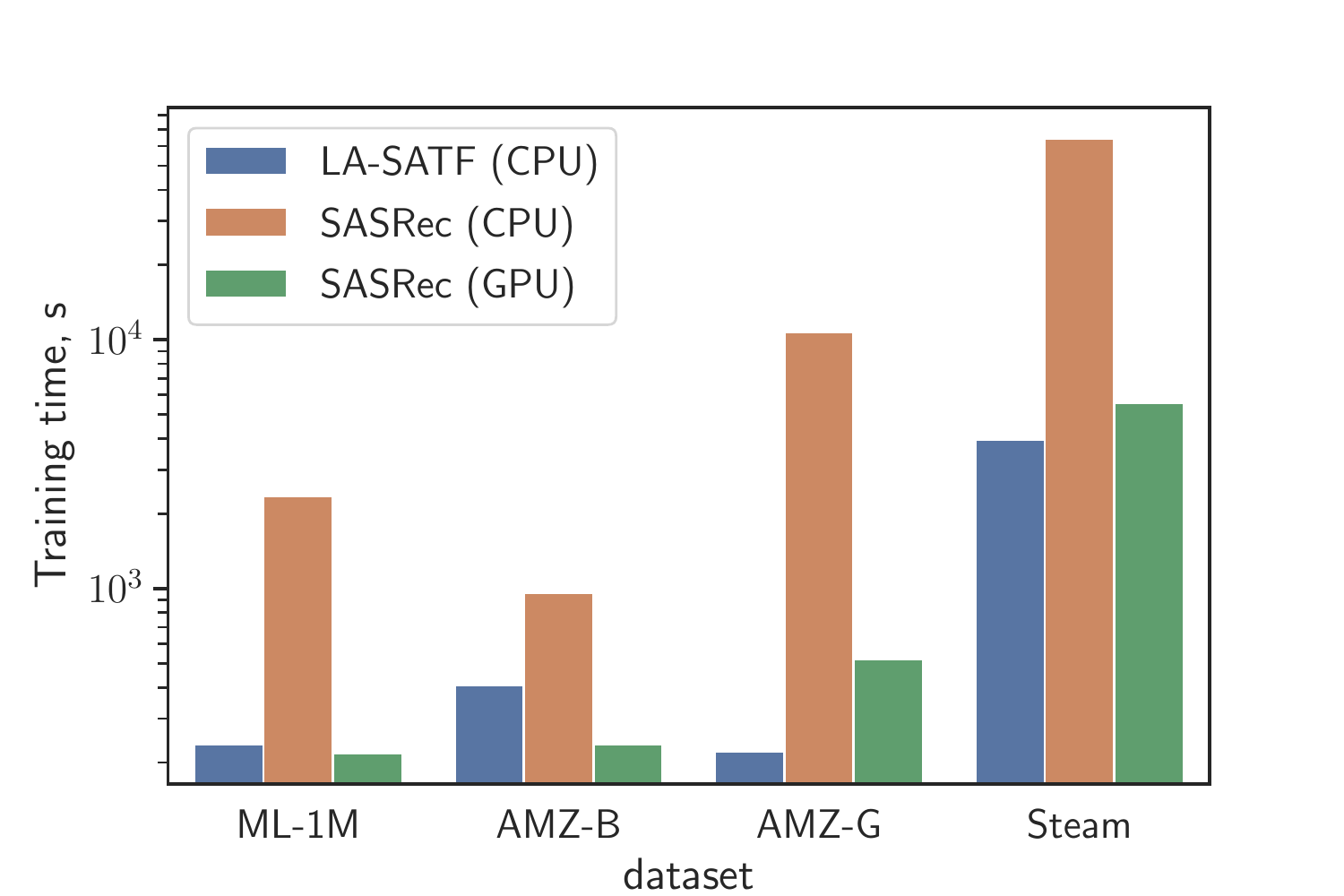}
  \caption{Training time corresponding to optimal configuration. Note, there are two types of measurements for the SASRec model: one CPU-based and another one GPU-based (marked with a dashed line).
  }
  \label{fig:optim_time}
\end{figure}

\subsection{Recommendations quality}
Overall, we observe a great level of parity between neural and our tensor-based attention approach.
In terms of both HR and NDCG metrics, there is a 50/50 distribution of best and second best scores between SASRec and our SA-SATF model.
In terms of the coverage metric, SASRec has a slight edge. However, it goes at the expense of the greater variability of this metric. For example, on the Steam dataset, it achieves several times lower coverage which means that SASRec's recommendations are not very personalized and are drawn from a small subset of popular items.

Interestingly, on the AMZ-B dataset, the LA-SATF model not only competes with SASRec, but even significantly outperforms it. We hypothesize that one of the reasons could be the fixed structure of the attention matrix $\matr{A}$, which in the case of this dataset may align well with the underlying mechanisms of user decision-making. Adaptively learning the weights as in the SASRec case may be less advantageous here. By looking at the reported in \cite[Fig. 4]{kang2018SARS} structure of the averaged attention weights learned by SASRec on this dataset, we see that only a few the most recent items matter for the next item prediction. Such structure is easily mimicked with the proposed banded form of $\matr{A}$, which in turn may provide additional robustness advantage for learning over noisy user behavior.

An opposite behavior is observed on the Steam dataset. The significant advantage in terms of the HR and NDCG metrics is now on the SASRec side. Remarkably, the GA-SATF model fails to learn meaningful patterns, as demonstrated by the very low HR and NDCG metrics. This may indicate that user decision-making processes have an intricate nature not captured by the fixed structure of the proposed attention mechanisms. Morever, by inspecting the ablation study provided by the authors of SASRec in \cite[Table~4]{kang2018SARS}, we note that only on this dataset the removal of positional embeddings actually improves the model. The adaptive self-attention mechanism turns out to be more advantageous here. However, as we mentioned earlier, on this dataset, SASRec tends to exploit trivial patterns with low recommendations diversity. The lack of diverse options may negatively impact user experience in practice. The LA-SATF model does a better job in this regard. Remarkably, the GA-SATF's global attention turns out to be incapable of improving any of the aspects of quality assessment in this case, which proves useful the implementation of the localized attention in LA-SATF. On the remaining two datasests, we observe that both SASRec and LA-SATF models exhibit a similar quality of predictions with the standard error range. The LA-SATF model has a nonsignificant advantage on ML-1M, while SASRec performs slightly but not significantly better on AMZ-G.

Surprisingly, the best performing model on AMZ-G is not sequential at all. Recall, SVD-based models are unrestricted and use the entire user history for generating predictions. It follows that on the AMZ-G dataset, knowing all user preferences provides more insights than sequential information. A possible remedy for LA-SATF would be to learn it incrementally as new data arrives in the system. That way, at every moment, the model would be updated with the most recent sequential information, but the old history would be implicitly encoded in the current model's latent space.

From practical viewpoints, searching for optimal values of hyper-parameters in new domains tends to be a bit easier with SASRec. The tensor-based approach may require more exploration for finding optimal ranges of values. Its hard-coded hyper-parameters of attention mechanism may significantly differ in different domains and thus may require an extensive search. On the other hand, fixed attention weights may provide more robustness for model training by preventing the model from focusing on outliers, which in turn is likely to positively affect the overall recommendations quality.

\subsection{Scalability}
As we mentioned earlier, there are certain obstacles that limit performing an in-depth computational performance comparison. PyTorch implementation of SASRec is generally optimized for GPU-based computations, while our solution is currently CPU-only. Nevertheless, the experimental setup described in Section~\ref{sec:scalability} still allows analyzing and comparing the general scalability trends of both approaches.

As shown in Fig~\ref{fig:timings}, both models exhibit similar asymptotic behavior. At larger values of item embeddings size, the asymptotic becomes linear in logarithmic scale, which corresponds well to the theoretical estimates provided in Section~\ref{sec:complexity}. Remarkably, when compared to the CPU-based run of SASRec, \emph{our approach shows more than an order of magnitude improvement} in terms of the training time. It is also interesting to note that for moderate dimensionality sizes, \emph{our tensor-based approach turns out to be even faster than the GPU-based neural counterpart}. This result indicates great potential for further adoption of our solution to GPU-based architectures, which may lead to considerable speedups over neural approach on all platforms.

We additionally measure an overall training time using the optimal configuration found during the hyper-parameters search phase. The results are show in Fig.~\ref{fig:optim_time}. We observe the same general trend as in the previous series of experiments: our tensor-based solution consistently outperforms its CPU-based neural counterpart on all datasets. In some cases it also becomes faster then the GPU-based implementation as well. The speedup is different depending on the dataset though, which is explained by differences in optimal hyper-parameter values that affect computational complexity, e.g., multilinear rank in LA-SATF model and the item embedding size or the number of attention blocks in SASRec.





%% file: text/conclusion.tex
\section{Conclusion}
We have proposed two sequential learning models based on the tensor factorization approach. These models enable slightly different attention mechanisms acting either on entire user sequences or within a context window of a fixed length typically much shorter than the sequence. The latter attention scheme proves to be more efficient in terms of the quality of recommendations and strongly competes with a more complicated deep learning self-attention. Our purely CPU-based implementation runs an order of magnitude faster than its neural competitor, and almost as fast as the latter's GPU-based version run on modern GPU.

The proposed approach is especially suitable in the domains where sequential information has a pronounced influence on the user decision-making process. The fixed structure of attention weights used by the tensor-based model helps to capture such sequential patterns efficiently. However, in more intricate cases with non-trivial dependencies, the adaptive self-attention mechanism of the transformer is likely to be more beneficial.

While our approach may not provide a universal solution, it is still a more lightweight alternative to existing state-of-the-art methods and can be advantageous in certain domains. It would be interesting to adapt the proposed approach to session-based and session-aware recommendation scenarios with repeating items in user sessions. Moreover, as there is no general assumption on the nature of data, it can be used in other disciplines where timeseries data is extremely incomplete. One such example is data from a network of sensors around the Earth that gather valuable information about climate and significantly depend on weather conditions, terrain, lighting, etc. The problem of missing data inevitably arises there.

Performance-wise, the use of standard and efficient optimization techniques in our tensor-based approach allows further improvements based on incremental learning schemes for handling online data streams. This may significantly extend the area of practical applications of the approach and seem to present another plausible direction for research.